%% file: main.tex
\documentclass[preprint,12pt]{elsarticle}

\usepackage[letterpaper,top=2cm,bottom=2cm,left=3cm,right=3cm,marginparwidth=1.75cm]{geometry}
\usepackage{amsmath}
\usepackage{array}
\usepackage{amssymb}
\usepackage{graphicx}
\journal{Computer Methods in Applied Mechanics and Engineering}
\begin{document}
\begin{frontmatter}

\title{
Physics-Informed Neural Networks and Radial Basis Functions for PDEs with Dirac Delta Sources
}
\tnotetext[t1]{This manuscript has been submitted to
Computer Methods in Applied Mechanics and Engineering for possible publication.}

\author[UIUC]{Manuel Reyna}
\author[UIUC]{Alexandre Tartakovsky\corref{cor1}}
\cortext[cor1]{Corresponding author: amt1998@illinois.edu}
\affiliation[UIUC]{
    organization={Department of Civil and Environmental Engineering, University of Illinois Urbana-Champaign}, 
    addressline={205 North Mathews Avenue}, 
    city={Urbana}, 
    postcode={61801}, 
    state={IL}, 
    country={USA}
}

\begin{abstract}
Physics-Informed Neural Networks (PINNs) are a machine learning method for solving forward and inverse Partial Differential Equations (PDEs). When applied to PDEs with Dirac delta functions in the forcing terms, boundary conditions, or initial conditions, PINNs require approximating them with smooth surrogate functions, a practice that can introduce significant modeling errors. In this work, we exploit the interpretation of PINNs as Residual Least Squares (RLS) methods and show that this perspective enables direct treatment of Dirac delta terms by integrating the weak-form equation. Among RLS formulations other than PINN, we focus on the Radial Basis Function (RBF) expansion (also known as a single-layer RBF Network). We show that while integrating out the Dirac delta in PINNs causes residuals to fail to converge to zero, RBF–RLS consistently provides good forward and inverse solutions to transport problems. We explain this finding using the Neural Tangent Kernel (NTK) theory. We test both approaches on linear PDEs that represent groundwater flow and transport in porous media and rivers. We solve inverse problems to fit synthetic data, noisy synthetic data, and real-world measurements.
\end{abstract}

\begin{highlights}

\item Dirac delta sources are included in the physics-informed loss by analytical integration
\item Radial basis functions accurately solve forward and inverse PDE problems
\item Deep neural networks fail to converge for PDEs with Dirac sources
\item Neural tangent kernel analysis explains the convergence failure

\end{highlights}

\begin{keyword}
Residual least squares\sep Dirac delta sources\sep Advection–diffusion equation\sep Radial basis functions\sep Physics-informed neural networks\sep Inverse problems\sep Transport in porous media \sep Groundwater flow \sep River solute transport 
\end{keyword}

\end{frontmatter}

\section{Introduction}

Advection--diffusion type equations with localized or instantaneous sources arise in a wide range of applications, including solute transport in porous media, groundwater flow, and tracer experiments in rivers. These processes are naturally modeled using Dirac delta distributions, either as forcing terms in the governing partial differential equation (PDE) or as impulsive initial or boundary conditions. Such formulations pose significant challenges due to singularities or undefined derivatives at some locations in the solutions. Additionally, in collocation methods (both exactly determined and overdetermined, i.e., solved by least squares optimization), the Dirac delta distribution cannot be sampled discretely with collocation points.

Among collocation methods, Physics-Informed Neural Networks (PINNs) have recently gained popularity for solving differential equations and for data assimilation in differential equation models \cite{raissi2019physics,karniadakis2021physics,cuomo}. PINNs use deep neural networks (DNNs) to approximate the state variables of the differential equation, and a least-squares collocation-like formulation to compute DNN parameters. PINNs have been applied to partial and ordinary differential equations describing various physical processes, including Navier-Stokes flow \cite{raissi2019physics,karniadakis2021physics}, Darcy flow \cite{tartakovsky2020physics}, transport problems described by the advection-dispersion equations (ADEs) \cite{he2021physics,Zhao2026Source}, to name a few.

Several approaches have been proposed to address Dirac delta sources in PINNs. Some studies integrate out the Dirac delta by using weak or variational formulations of the governing equations \cite{huan_2021,e_yu_2018,Wang2025fracture,Xu2020hybrid}. Other works rely on partial analytical solutions involving Green's functions to explicitly account for point sources \cite{park_jo_2024}. In practice, however, the most common approach in PINNs is to approximate Dirac delta distributions with a mollifier, e.g., Gaussian functions \cite{rosa_pompameo_litvinenko_cuomo_2025,huan_2021,Kapoor_2024,Liu2024,Zhao2026Source}. The same has been done in other collocation methods \cite{suarez_jacobs_2016}. While convenient, this approximation introduces an additional modeling parameter and can lead to significant errors.

Residual Least Squares (RLS) methods have a long history in the numerical solution of differential equations across many disciplines \cite{eason_review_1976}. They arise naturally in data assimilation problems due to the ubiquity of least squares regression as a framework for fitting noisy observations. RLS methods admit multiple equivalent mathematical interpretations. They can be viewed as weighted residual methods with weighting functions of the form $\partial R / \partial \theta_i$ (forming a weak form equation), where $R$ denotes the strong-form residual of the differential equation and $\theta_i$ are the parameters of the solution approximation \cite{kharazmi_hp-vpinns_2021}. Alternatively, they can be interpreted as overdetermined collocation methods solved through least squares optimization \cite{hanke_convergence_2021}. As a result, RLS methods share foundational elements with Galerkin methods and regularized regression techniques.

Since Dirac delta sources can be integrated out using weak formulations of the differential equations \cite{reddy1993introduction,e_yu_2018,Xu2020hybrid}, and PINNs can be thought of as an RLS method with discretized integrals (if the distribution and weighting of residual points are appropriate), PINNs can be reformulated for general PDE problems such that Dirac delta sources need not be approximated with mollifiers. In this work, we find that although this approach can yield good results with deep neural networks, some losses tend to stagnate, limiting accuracy. Specifically, residuals of the form $r=\mathcal{L}(\hat{u})$, where $\mathcal{L}$ is the differential operator defining the PDE or its boundary or initial conditions and $\hat{u}$ is the approximation function (in this case, the Neural Network), do not tend to $0$ in training even when the Neural Network can perfectly approximate the solution (in the limit of an infinitely wide network), unlike in typical formulations of PINNs \cite{Wang2022NTK}. As supported by Neural Tangent Kernel (NTK) theory and numerical tests, residuals at the location of the point source are coupled with residuals at locations within a radius of influence, which cannot be reduced by increasing the number of trainable parameters. Thus, we find that errors in PINNs with integrated Dirac deltas cannot be reduced by increasing the complexity of the Neural Network.

Due to these limitations of PINNs, we focus on an alternative RLS formulation using Radial Basis Function (RBF) expansions to approximate the solution of the governing PDE. The use of the RBF expansion as a function approximator is sometimes described as single-layer RBF networks, with most applications corresponding to purely data-driven problems \cite{lowe1988multivariable,schwenker_three_2001,Ghosh2001}. RBF methods have long been used for the numerical solution of differential equations, most notably through the Kansa collocation method, in which the strong-form residual is enforced at the RBF centroids \cite{kansa_multiquadricsscattered_1990_1,kansa_multiquadricsscattered_1990_2}. Here, we adopt a different approach by fitting the RBF approximation by minimizing the squared residual integrated over the space--time domain, yielding an RLS formulation. Unlike standard RBF collocation methods, we use RBFs to represent both spatial and temporal dimensions. While multiple works have focused on the solution of forward problems using RBF–RLS methods \cite{kwok_convergence_2009,Alqezweeni2018Efficient,bai_physics-informed_2023}, the only case of its use to solve inverse problems, to our knowledge, was an application to estimate the coefficients in the RBF expansions of both the unknown function state and the parameter space in elliptic PDEs \cite{gorbachenko_application_2017} \cite{gorbachenko_solving_2017} \cite{gorbachenko_physics-informed_2023}. In RBF–RLS, with fixed centroids and shape parameters, the coefficients of the approximation function (which constitute the totality of the learned parameters) are linear. Then, the weak-form equation can be transformed into a system of algebraic equations by setting its gradient to zero and computing integrals that are independent of the coefficients. Crucially for the solution of problems with Dirac delta sources, the radius of influence of many RBFs can be made arbitrarily small as the number of centroids increases. This reduces the coupling between the residuals at the source location and at other locations, allowing the residuals to converge to zero in the limit of an infinite expansion.

Section \ref{sec:methods}, 
 formulates the residual least-squares method for a generic PDE problem, explains how Dirac delta point sources are integrated out, derives a discrete approximation of the weak-form residuals, introduces the RBF expansion within the RLS framework, and describes the parameter-estimation procedure.
In Section \ref{sec:num_exp}, we present four different problems on which we test our methods to obtain forward and inverse solutions using RBF–RLS and comparing to PINNs, when PINN solutions are adequate. We choose these problems to represent different types of transport equations and Dirac delta sources. Finally, we analyze the behavior of the PINN losses using Neural Tangent Kernel (NTK) theory in Section \ref{sec:PINN-failure} and give general conclusions in Section \ref{sec:conclusions}.

\section{Methods}\label{sec:methods}
In this section, we first introduce the RLS method for PDEs with a Dirac delta source term (Section \ref{sec:RLS}), and its discrete formulation. The discrete formulation of the RLS method for PDEs with point-source terms is a novel contribution of this paper and can be used with different approximations of the state variable, including DNNs (as in the PINN method) and RBFs (as in the proposed RBF–RLS method).  In Section \ref{sec:RBF-expansion}, we review the RBF approximation of functions. In Sections \ref{sec:methods-forward} and \ref{sec:Inverse_RBF}, we formulate the RBF–RLS method for forward and inverse problems, including Dirac delta sources, respectively, which are the main contributions of this paper. 

\subsection{Residual Least Squares Method with Dirac delta sources}\label{sec:RLS}

To introduce the RLS method, consider the PDE problem
\begin{align}\label{eq:general PDE}
\mathcal{L}(u(\mathbf{x}, t); \mathbf{v}) &= f(\mathbf{x},t;\mathbf{v}), \mathbf{x} \in \Omega, t\in[0,T]
\\
\mathcal{B}(u(\mathbf{x}, t); \mathbf{v}) &= g(\mathbf{x}, t;\mathbf{v}), \mathbf{x} \in \partial\Omega\\
u(\mathbf{x}, t = t_0) &= U(\mathbf{x};\mathbf{v}),
\label{eq:ITC}
\end{align}
where $\mathcal{L}$ is a differential operator, $\mathcal{B}$ is a boundary condition operator, $\mathbf{x}$ is the spatial coordinate vector, $t$ is time, $\Omega$ is a bounded domain in $R^n$ with the boundary  $\partial\Omega$, $u(\mathbf{x},t)$ is the state variable, $\mathbf{v}$ is a vector of system parameters, and $f(\mathbf{x}, t)$, $g(\mathbf{x},t)$, and $U(\mathbf{x})$ are known functions. 

We consider the case with a point source in the forcing of the PDE, i.e., 
\begin{equation}\label{eq:f_Dirac}
f=\hat{f}(\mathbf{x},t) + M^{\text{ep}}(t) \delta(\mathbf{x}-\mathbf{x}^{\text{ep}}),
\end{equation}
where $\hat{f}(\mathbf{x},t)$ is the dispersed source term, $\delta(x)$ is the Dirac delta distribution, and $M^{\text{ep}}(t)$ and $\mathbf{x}^{\text{ep}}\in \Omega$ represent the mass flux and location of the concentrated (point) source term. A more general formulation, with combinations of many Dirac delta functions in the PDE forcing or in the boundary or initial conditions, is given in \ref{app:general_dirac_delta}. 

The solution to a PDE can be found by approximating $u$ with a parameterized function $\hat{u}(\mathbf{x},t;\boldsymbol{\theta})$ and computing the parameters $\boldsymbol{\theta}$ by solving the residual least-squares minimization problem \cite{eason_review_1976,weighted2014}
\begin{align} \label{eq:LLR-complete}
    \boldsymbol{\theta}^{\text{opt}} 
    = &\min_{\boldsymbol{\theta}} \Big[\mathcal{R} (\boldsymbol{\theta}) \Big] = \min_{\boldsymbol{\theta}} \Big[ \lambda^{\text{e}} \mathcal{R}^{\text{e}}(\boldsymbol{\theta}) + \lambda^{\text{b}} \mathcal{R}^{\text{b}}(\boldsymbol{\theta}) + \lambda^{\text{i}} \mathcal{R}^{\text{i}}  (\boldsymbol{\theta}) \Big]
    \\\nonumber
    =&
    \min_{\boldsymbol{\theta}} \Big[
    \frac{\lambda^{\text{e}}}{||\Omega|| T }\int_\Omega \int_0^T  [\mathcal{L}(\hat{u})-f]^2 dt d\mathbf{x} + \frac{\lambda^{\text{b}}}{||\partial \Omega || T } \int_{\partial \Omega} \int_0^T  [\mathcal{B}(\hat{u}) - g]^2 dt d\mathbf{x} \\ \nonumber
    &+ \frac{\lambda^{\text{i}}}{||\Omega|| }\int_\Omega  [\hat{u}(t=t_0) - U ]^2  d\mathbf{x}  \Big],
    \\\nonumber
    =&
    \min_{\boldsymbol{\theta}} \Big[
    \frac{\lambda^{\text{e}}}{||\Omega||T } \left(\int_\Omega \int_0^T  [\mathcal{L}(\hat{u})^2 - 2\mathcal{L}(\hat{u})\hat{f}]dt d\mathbf{x}  -2\int_0^T  M^{\text{ep}}(t)\mathcal{L}(\hat{u}(\mathbf{x}^{\text{ep}}))  dt \right.
    \\\nonumber
    &\left.+ 2\int_0^T  [\hat{f}(\mathbf{x}^{\text{ep}})M^{\text{ep}}(t) ]dt  +\int_\Omega \int_0^T   \hat{f}^2dt d\mathbf{x} +  \int_\Omega \int_0^T  [ M^{\text{ep}}(t) \delta(\mathbf{x}-\mathbf{x}^{\text{ep}})]^2dt d\mathbf{x}\right) \\ \nonumber
    &+\frac{\lambda^{\text{b}}}{||\partial \Omega || T } \int_{\partial \Omega} \int_0^T  [\mathcal{B}(\hat{u}) - g]^2 dt d\mathbf{x} + \frac{\lambda^{\text{i}}}{||\Omega|| }\int_\Omega  [\hat{u}(t=t_0) - U ]^2  d\mathbf{x}  \Big].
\end{align}
where $\lambda^{\text{e}}$, $\lambda^{\text{b}}$, and $\lambda^{\text{i}}$ are the squared-residual weights of the PDE, the boundary conditions, and the initial condition, respectively. The corresponding loss terms are $\mathcal{R}^{\text{e}}$, $\mathcal{R}^{\text{b}}$, and $\mathcal{R}^{\text{i}}$, and $\mathcal{R}$ is the total loss function.

Since the term $2\int_0^T  [\hat{f}(\mathbf{x}^{\text{ep}})M^{\text{ep}}(t) ]dt  +\int_\Omega \int_0^T   \hat{f}^2dt d\mathbf{x} $ is independent of $\boldsymbol{\theta}$, it can be eliminated from the minimization objective. Based on the properties of the Dirac delta distribution, the term $\int_\Omega \int_0^T  [ M^{\text{ep}}(t) \delta(\mathbf{x}-\mathbf{x}^{\text{ep}})]^2dt d\mathbf{x}$ is undetermined. However, we can treat the Dirac delta as the limit of the Gaussian function: $\delta(x)=\lim_{\epsilon \to 0} \frac{\exp(-x^2/(2\epsilon^2))}{\sqrt{2\pi \epsilon^2}}$. For any arbitrarily small $\epsilon$, the term $\int_\Omega \int_0^T  [ M^{\text{ep}}(t) \delta(\mathbf{x}-\mathbf{x}^{\text{ep}})]^2dt d\mathbf{x}$ is finite and independent of $\boldsymbol{\theta}$; therefore, it can also be eliminated from the minimization problem without affecting its solution. Then, Eq. \eqref{eq:LLR-complete} can be reduced to:
\begin{align} \label{eq:LLR}
    \boldsymbol{\theta}^{\text{opt}} 
    =&
    \min_{\boldsymbol{\theta}} \Big[
    \frac{\lambda^{\text{e}}}{||\Omega|| T} \left( \int_\Omega \int_0^T  [\mathcal{L}(\hat{u})^2 - 2\mathcal{L}(\hat{u})\hat{f}]dt d\mathbf{x}  -2\int_0^T  M^{\text{ep}}(t)\mathcal{L}(\hat{u}(\mathbf{x}^{\text{ep}}))  dt \right.
    \\\nonumber
    &\qquad+ \frac{\lambda^{\text{b}}}{||\partial \Omega || T } \int_{\partial \Omega} \int_0^T  [\mathcal{B}(\hat{u}) - g]^2 dt d\mathbf{x} + \frac{\lambda^{\text{i}}}{||\Omega|| }\int_\Omega  [\hat{u}(t=t_0) - U ]^2  d\mathbf{x}  \Big].
\end{align}

Solutions of the minimization problem in Eq. \eqref{eq:LLR} satisfy the zero gradient condition
\begin{align} \label{eq:LLR-derivatives}
    \mathbf{0} =& 
    \frac{2\lambda^{\text{e}}}{||\Omega||T }\left(\int_\Omega \int_0^T  [\mathcal{L}(\hat{u})-\hat{f}]\frac{\partial \mathcal{L}(\hat{u})}{\partial \boldsymbol{\boldsymbol{\theta}}} dt d\mathbf{x}  - \int_0^T  M^{\text{ep}}(t)  \frac{\partial \mathcal{L}(\hat{u}(\mathbf{x}^{\text{ep}}))}{\partial \boldsymbol{\boldsymbol{\theta}}} dt\right)
    \\ \nonumber
    &+ \frac{2\lambda^{\text{b}}}{||\partial \Omega || T } \int_{\partial \Omega} \int_0^T  [\mathcal{B}(\hat{u}) - g] \frac{\partial \mathcal{B}(\hat{u}) }{\partial \boldsymbol{\boldsymbol{\theta}}} dt d\mathbf{x} + \frac{2\lambda^{\text{i}}}{||\Omega|| }\int_\Omega  [\hat{u}(t=t_0) - U ] \frac{\partial \hat{u}(t=t_0)}{\partial \boldsymbol{\boldsymbol{\theta}}}  d\mathbf{x}.
\end{align}
This equation is commonly used to solve linear least-squares problems in which $\hat{u}$ is linear in $\boldsymbol{\theta}$ and the PDE is linear in $u$. 
The integrals containing $\int_\Omega \int_0^T  \mathcal{L}(\hat{u})\frac{\partial \mathcal{L}(\hat{u})}{\partial \boldsymbol{\boldsymbol{\theta}}} dt d\mathbf{x}$ have the form of a weighted residual method where the strong form residual of the differential equation $\mathcal{L}(\hat{u})$ is weighted by its derivatives over the coefficients of the expansion $\frac{ \partial \mathcal{L}(\hat{u})}{\partial \boldsymbol{\boldsymbol{\theta}}}$ \cite{nguyen_space-time_1984}. Since the number of derivatives is equal to the number of unknowns, the resulting system of equations is perfectly determined, and, if well posed, should have a finite number of solutions. 

Various representations $\hat{u}(\mathbf{x},t;\boldsymbol{\theta})$ could be used, including a combination of Lagrange polynomials as in the least-squares finite element (LSFE) method \cite{nguyen1984space}, DNNs in the standard PINN method (where  $\boldsymbol{\theta}$ is the assembly of weights and biases of the DNN), and KL expansion in the physics-informed conditional Karhunen–Loève (PICKLE) method \cite{TARTAKOVSKY2021109904}. 

In this work, we use DNNs to approximate the PDE solution, yielding the PINN method, which we formulate for PDEs with point sources using the loss in Eq. \eqref{eq:LLR}. We also approximate solutions of PDEs with point sources using RBFs, which we refer to as the RBF–RLS methods. We compare the two methods for linear PDEs.    
The RBF expansion is linear in $\boldsymbol{\theta}$. Therefore, the RBF solution can be obtained using Eq. \eqref{eq:LLR-derivatives}.  
For nonlinear PDE problems, RBF–RLS solutions can be obtained using Eq. \eqref{eq:LLR} \cite{eason_review_1976}.

\subsubsection{Discrete formulation}

For many types of equations and functional approximations of state variables, the integrals in Eqs. \eqref{eq:LLR} and \eqref{eq:LLR-derivatives} are either impossible or impractical to compute analytically, and they must be computed with quadrature methods. Even when analytical integration is possible, as in the RBF–RLS method for linear PDEs, we also use quadrature methods because of their ease of implementation and negligible quadrature errors with a very moderate number of quadrature points. The discrete forms of Eqs. \eqref{eq:LLR} and \eqref{eq:LLR-derivatives}  are 
\begin{align} \label{eq:DLLR}
\boldsymbol{\theta}^{\text{opt}} = \min_{\boldsymbol{\theta}} \Big[&
 \lambda^{\text{e}}\left(\sum_i^{N_{\Omega\times T}^\text{e}} w_{i} [\mathcal{L}(\hat{u}_{i}^\text{e})^2-2 \mathcal{L}(\hat{u}_{i}^\text{e})\hat{f}_{i}]-2 \sum_i^{N_{T}^\text{ep}} w_{i}' M^{\text{ep}}(t_i^\text{ep}) \mathcal{L}(\hat{u}^{\text{ep}}_i)\right)
 \\\nonumber
 &+ \lambda^{\text{b}}\sum_i^{N_{\partial\Omega\times T}^\text{b}} w''_i [\mathcal{B}(\hat{u}_i^{\text{b}}) -g_i]^2  + \lambda^{\text{i}}\sum_i^{N_{\Omega}^\text{i}} w'''_{i} [\hat{u}_{0,i}^\text{i} - U_i ]^2 \Big],
\end{align}
and
\begin{align} \label{eq:DLLR-derivatives}
\mathbf{0} = &
 2\lambda^{\text{e}}\left(\sum_i^{N_{\Omega\times T}^\text{e}}w_i [\mathcal{L}(\hat{u}_i^{\text{e}})-\hat{f}_i] \frac{\partial \mathcal{L}(\hat{u}_i^{\text{e}})}{\partial \boldsymbol{\boldsymbol{\theta}}}- \sum_i^{N_{T}^{\text{ep}}} w_i' M^{\text{ep}}(t_i^\text{ep}) \frac{\partial \mathcal{L}(\hat{u}_i^{\text{ep}})}{\partial \boldsymbol{\boldsymbol{\theta}}}\right)
 \\\nonumber
 &+ 2\lambda^{\text{b}}\sum_i^{N_{\partial\Omega\times T}^\text{b}} w''_i [\mathcal{B}(\hat{u}_i^{\text{b}}) -g_i]\frac{\partial \mathcal{B}(\hat{u}_i^{\text{b}}) }{\partial \boldsymbol{\boldsymbol{\theta}}}  + 2\lambda^{\text{i}}\sum_i^{N_{\Omega}^\text{i}} w'''_{i} [\hat{u}_{0,i}^\text{i} - U_i ] \frac{\partial \hat{u}_{0,i}^\text{i}}{\partial \boldsymbol{\boldsymbol{\theta}}} 
\end{align}
where $\hat{u}_i^{\text{e}}=\hat{u}(\mathbf{x}_i^{\text{e}},t_i^{\text{e}};\boldsymbol{\theta})$, $\hat{u}_i^{\text{b}}=\hat{u}(\mathbf{x}_i^{\text{b}},t_i^{\text{b}};\boldsymbol{\theta})$, $\hat{u}^{\text{ep}}_i=\hat{u}(\mathbf{x}^{\text{ep}},t_i^\text{ep})$, $\hat{u}_{0,i}^\text{i} = \hat{u} (\mathbf{x}_i^\text{i},t_0;\boldsymbol{\theta})$, $ \hat{f}_i = f(\mathbf{x}_i^{\text{e}},t_i^{\text{e}})$, $g_i = g(\mathbf{x}_i^{\text{b}},t_i^{\text{b}})$, and $U_i = U(\mathbf{x}_i^{\text{i}})$.  The coefficients are $w_i = \frac{\hat{w}_i}{N_{\Omega\times T}^\text{e}}$, $w'_i = \frac{\hat{w}'_i}{ N_{T}^\text{ep} ||\Omega||}$, $w_i'' = \frac{ \hat{w}_i''}{N_{\partial\Omega\times T}^\text{b}}$ and $w_{i}'''=\frac{\hat{w}_{i}''' }{N_{\Omega}^\text{i}} $, where $N_{\Omega\times T}^\text{e}$, $N_{T}^\text{ep}$, $N_{\partial\Omega\times T}^\text{b}$, and $N_{\Omega}^\text{i}$ are the number of quadrature points in the PDE residual ($\mathbf{x}_i^{\text{e}},t_i^{\text{e}}$), in the Dirac delta residual ($t_i^{\text{ep}}$),  in the boundary condition residual ($\mathbf{x}_i^{\text{b}},t_i^{\text{b}}$), and in the initial condition residual ($\mathbf{x}_i^{\text{i}}$), respectively. $\hat{w}_i$, $ \hat{w}_i'$, $\hat{w}_{i}''$, and $\hat{w}_{i}'''$ are quadrature weights. In RBF–RLS, we use the trapezoid rule for integration with equally spaced centroids. For PINN, we use Monte Carlo integration with a uniform probability distribution (for boundary and initial conditions) and Latin Hypercube sampling (for the PDE), leading to all weights being equal to $1$. Note that PINNs can be considered only a special case of the RLS method if the distribution of residual points and the weighting of the losses yield appropriate quadratures of the loss terms.

The discrete formulation of the PDE solutions with a Dirac delta function in $g$ and/or $U$ can be obtained similarly by discretizing the corresponding integral equations in \ref{app:general_dirac_delta}.

\subsection{Radial Basis Functions expansion}\label{sec:RBF-expansion}

As is done in various numerical methods, ranging from finite elements to smoothed particle hydrodynamics, RBF–RLS uses a linear combination of basis functions to approximate the solution to the PDE. Radial Basis Functions (RBFs), i.e., functions that depend only on the distance between any point and a centroid, are often used in Smoothed Particle Hydrodynamics (SPH) and Kansa methods. However, unlike those methods where the basis functions are functions of space only and the coefficients are functions of time, here we use RBFs to construct basis functions $\phi_k(\mathbf{x},t)=\phi(||\mathbf{x}-\mathbf{x}^c_k||,|t-t^c_k|)$ in space--time, yielding the approximation of a function $u(\mathbf{x},t)$ in terms of unknown (but constant in space and time) coefficients $u_k$
\begin{align}\label{eq:RBF-Disc-Approximation}
   {u}(\mathbf{x},t) \approx \hat{u}(\mathbf{x},t;\mathbf{u}) =& \sum_{k=1}^N u_k \phi(||\mathbf{x}-\mathbf{x}^c_k||,|t-t^c_k|)
   \\\nonumber
   =&\sum_{k=1}^N u_k W(||\mathbf{x}-\mathbf{x}^c_k||,h_x) W(t-t^c_k,h_t)  ,
\end{align}
where $W(||\mathbf{x}-\mathbf{x}^c_k||,h_x)$ and $W(t-t^c_k,h_t)$ are RBFs in space and time, respectively, with fixed smoothing lengths or shape parameters $h_x$ and $h_t$, and centroids $\mathbf{x}^c_k$ and $t^c_k$. The vector $\mathbf{u}=[u_1,...,u_N]^\mathsf{T} $ groups all RBF coefficients. 

Different forms of RBFs can be used  \cite{mai-duy-approximation-2003}. In our work, we assume that $W$ is the Gaussian function $W(z-z^c_k,h)=\frac{1}{\sqrt{2\pi}h} \exp \left(-\frac{(z-z^c_k)^2}{2h^2}\right)$. The RBF–RLS method is local in the sense that points far apart interact little, even when squared residuals are minimized globally in Eq. \eqref{eq:LLR}. Although some works set the shape parameter \cite{Alqezweeni2018Efficient,bai_physics-informed_2023} and the centroid locations \cite{Alqezweeni2018Efficient} as trainable parameters, we fix them to ensure the approximation function is linear in the parameters.

\subsection{The RBF–RLS method for forward problems}\label{sec:methods-forward}

An RLS solution to the PDE \eqref{eq:general PDE}-\eqref{eq:ITC} approximated by an RBF in Eq. \eqref{eq:RBF-Disc-Approximation} is obtained from Eq. \eqref{eq:LLR}, where the set of parameters $\boldsymbol{\theta}$ is the vector of coefficients $\mathbf{u}$. The optimal set of parameters $\mathbf{u}$ satisfies Eq. \eqref{eq:LLR-derivatives} or its discrete approximation in Eq. \eqref{eq:DLLR-derivatives}. One advantage of using a linear approximation function is that, if the operators $\mathcal{L}$ and $\mathcal{B}$ are polynomials of $u$ and its derivatives, Eq. \eqref{eq:LLR-derivatives}  can be written as a system of polynomial equations of  $\mathbf{u}$ with constant coefficients. If the operators are linear,  this system of equations is linear and has the form:
\begin{align}\label{eq:linear_weak_form}
    A\mathbf{u}=&\mathbf{b},
    \\
    \label{eq:linear_eq_RBF_def_A}
    A_{kl} =& \frac{\lambda^{\text{e}}\int_\Omega \int_0^T   \mathcal{L}(\phi^k) \mathcal{L}(\phi^l) dt d\mathbf{x}}{||\Omega|| T }  + \frac{\lambda^{\text{b}}\int_{\partial \Omega} \int_0^T  \mathcal{B}(\phi^k)  \mathcal{B}(\phi^l) dt d\mathbf{x}}{||\partial \Omega || T }  + \frac{\lambda^{\text{i}} \int_\Omega  \phi_{0}^k \phi_{0}^l  d\mathbf{x}}{||\Omega|| },
    \\\label{eq:linear_eq_RBF_def_b}
    b_{k} =&\frac{\lambda^{\text{e}}}{||\Omega || T } \left(\int_{\Omega} \int_0^T   \hat{f} \mathcal{L}(\phi^k) dt d\mathbf{x} + \int_0^T   M^{\text{ep}} \mathcal{L}(\phi^k_{\text{ep}}) dt \right)  
    \\\nonumber
    &+\frac{\lambda^{\text{b}} \int_{\partial \Omega} \int_0^T   g \mathcal{B}(\phi^k) dt d\mathbf{x}}{||\partial \Omega || T } + \frac{\lambda^{\text{i}} \int_\Omega   U \phi_{0}^k d\mathbf{x} }{||\Omega|| },
\end{align}
where $\mathcal{L}(\phi^{k})$ and $\mathcal{B}(\phi^k)$ are the differential operators applied to the basis functions $\phi^k=\phi(||\mathbf{x}-\mathbf{x}^c_k||,|t-t^c_k|)$, $\phi^k_{\text{ep}}=\phi(||\mathbf{x}^{\text{ep}}-\mathbf{x}^c_k||,|t-t^c_k|)$, and $\phi_{0}^k =\phi(||\mathbf{x}-\mathbf{x}^c_k||,|t_0-t^c_k|)$.

Eq. \eqref{eq:linear_eq_RBF_def_A} has a unique solution. On the other hand, a DNN approximation of $u(\mathbf{x},t)$ in Eq. \eqref{eq:DLLR} or \eqref{eq:LLR} results in a non-linear least-squares problem, which takes longer to solve and may not yield unique solutions.

\subsection{Data assimilation and parameter estimation}\label{sec:Inverse_RBF}

In PINN, inverse solutions are obtained by jointly minimizing over the Neural Network weights $\boldsymbol{\theta}$ and the system parameters $\mathbf{v}$, while adding a data-squared residual to the physics squared residuals.  
Specifically, given $N^*$ measurements $u^*_m$ of the state $u$ at space--time locations $(\mathbf{x}^*_m,t^*_m)$,  the PINN inverse solution is defined as
\begin{equation}\label{eq:PINN_inverse}
    \boldsymbol{\theta}^{\text{opt}},\mathbf{v}^{\text{opt}} 
    = \min_{\boldsymbol{\theta,\mathbf{v}}} \left[\mathcal{R} (\boldsymbol{\theta},\mathbf{v}) + \frac{\lambda^*}{N^*}\sum_{m=1}^{N^*} [\hat{u}(\mathbf{x}^*_m,t^*_m;\boldsymbol{\theta}) - u^{*}_m]^2 \right].
\end{equation}

For RBF–RLS, we instead formulate the inverse solution as a minimization over system parameters $\mathbf{v}$ of the data mismatch in the forward solution, an approach particularly useful for linear PDEs, where the forward solution is obtained quickly by solving a linear problem. We define $\mathbf{u}^{\text{opt}}(\mathbf{v})$ as the solution of the forward RBF–RLS problem for arbitrary $\mathbf{v}$. For a linear problem, $\mathbf{u}^{\text{opt}}(\mathbf{v})= A(\mathbf{v})^{-1}\mathbf{b}(\mathbf{v})$, as given by Eq. \eqref{eq:linear_eq_RBF_def_A}. The inverse problem can be formulated generally (not only for linear problems) as
\begin{equation}\label{eq:RBF_inverse}
    \mathbf{v}^* = \min_{\mathbf{v}} \left[R^*(\mathbf{v})\right] ,
\end{equation}
where 
\begin{equation}
    R^*(\mathbf{v})=\sum_{m=1}^{N_m} [\hat{u}(\mathbf{x}^*_m,t^*_m;\mathbf{u}^{\text{opt}}(\mathbf{v})) - u^{*}_m]^2=\mathbf{e}^\mathsf{T} \mathbf{e}=(\Phi^* \mathbf{u}^{\text{opt}}(\mathbf{v})-\mathbf{u}^{*})^\mathsf{T}  (\Phi^* \mathbf{u}^{\text{opt}}(\mathbf{v})-\mathbf{u}^{*})
\end{equation}
and the matrix $\Phi^*$ is formed by elements $\Phi^*_{km}=\phi(||\mathbf{x}_m^*-\mathbf{x}^c_k||,|t_m^*-t^c_k|)$, and $\mathbf{e}=\Phi^* \mathbf{u}^{\text{opt}}(\mathbf{v})-\mathbf{u}^{*}$. 

To solve the minimization problem in Eq. \eqref{eq:RBF_inverse} and estimate parameters, we use the methods \textit{Nelder-Mead}, \textit{L-BFGS-B}, and \textit{trust-exact} from \textit{scipy.minimize} \cite{virtanen2020scipy}. We use these to compare the cases where an analytical gradient for $R^*(\mathbf{v})$ is computed, where an analytical gradient is computed but not a Hessian, and where both are computed. We found those methods to be optimal for our unconstrained optimization problem. This comparison is done to evaluate the trade-off between the computational cost of computing gradients and Hessians and the improved convergence of optimization methods that leverage them.

The gradient and Hessian of the objective function are
\begin{equation}
    \frac{\partial R^*(\mathbf{v})}{\partial v_{\alpha}} = 2 \left(\mathbf{e}^\mathsf{T}  \frac{\partial \mathbf{e}}{\partial v_{\alpha}}\right),\quad \frac{\partial^2 R^*(\mathbf{v})}{\partial v_{\alpha} \partial v_{\beta}}=2\left(\frac{\partial \mathbf{e}}{\partial v_{\alpha}}^\mathsf{T}  \frac{\partial \mathbf{e}}{\partial v_{\beta}}+\mathbf{e}^\mathsf{T}  \frac{\partial^2 \mathbf{e}}{\partial v_{\alpha}\partial v_{\beta}}\right),
\end{equation}
with $\frac{\partial \mathbf{e}}{\partial v_{\alpha}} = \Phi^* \frac{\partial \mathbf{u}^{\text{opt}}}{\partial v_{\alpha}}$, and $\frac{\partial^2 \mathbf{e}}{\partial v_{\alpha}\partial v_{\beta}} = \Phi^* \frac{\partial^2 \mathbf{u}^{\text{opt}}}{\partial v_{\alpha}\partial v_{\beta}}$. For linear problems, the gradient and Hessian of $\mathbf{u}^{\text{opt}}$ are obtained by differentiating Eq. \eqref{eq:linear_weak_form}
\begin{equation}\label{eq:grad_hess_linear_system}
    A \frac{\partial \mathbf{u}}{\partial v_{\alpha}} = \frac{\partial \mathbf{b}}{\partial v_{\alpha}} - \frac{\partial A}{\partial v_{\alpha}} \mathbf{u},\quad A \frac{\partial^2 \mathbf{u}}{\partial v_{\alpha}\partial v_{\beta}} = \frac{\partial^2 \mathbf{b}}{\partial v_{\alpha}\partial v_{\beta}} - \frac{\partial^2 A}{\partial v_{\alpha}\partial v_{\beta}} \mathbf{u} - \frac{\partial A}{\partial v_{\alpha}} \frac{\partial \mathbf{u}}{\partial v_{\beta}}- \frac{\partial A}{\partial v_{\beta}} \frac{\partial \mathbf{u}}{\partial v_{\alpha}}.
\end{equation}

Each matrix solution in Eqs. \eqref{eq:linear_weak_form} and \eqref{eq:grad_hess_linear_system} take up most of the computational time of inverse solutions. Since all the systems have the same matrix $A$, we solve them using a single Cholesky decomposition to accelerate the code. For $\mathcal{L}$ and $\mathcal{B}$ linear in $\mathbf{v}$ and $\mathbf{u}$ as well as $f$, $g$, and $h$ linear in $\mathbf{v}$, the components of the matrix $A(\mathbf{v})$, the vector $\mathbf{b}(\mathbf{v})$, and their derivatives are given in \ref{sec:coefficients}. All the problems we analyze involve operators that are linear in the parameters and state, but involve a product of the parameters and state.

\section{Numerical experiments}\label{sec:num_exp}

We consider four time-dependent one-dimensional PDE problems, including (1) Advection Dispersion Equation (ADE) with a point source initial condition, (2) ADE with a step boundary condition,  (3) ADE with first-order mobile-immobile exchange and a point source initial condition, and (4) Diffusion Equation (DE) with point source forcing. All these problems are defined on the normalized space--time domain $(x,t) \in [0,1]\times(0,1]$. For all PDEs, we obtain forward solutions and inverse solutions using synthetic or experimental measurements. 

For Problems 1, 2, and 4, we evaluate the errors on a uniform internal $n_x \times n_t$ space-time grid, where we set $n_x = n_t = 200$. Additional equally spaced grid points are placed along the boundaries of the space-time domain. We estimate the accuracy of the methods using  ``global'' Root Mean Square Error (RMSE) over the space--time domain
\begin{equation}
    \varepsilon = \sqrt{\frac{\sum_{i=1}^{n_x}\sum_{j=1}^{n_t} \left(\hat{u}(x_i,t_j)-u(x_i,t_j)\right)^2 }{n_x n_t}},
\end{equation}
and time-dependent and space-dependent RMSEs (averaged over the space domain and time domain, respectively):
\begin{equation}
    \varepsilon_x(t) = \sqrt{\frac{\sum_{i=1}^{n_x} \left(\hat{u}(x_i,t)-u(x_i,t)\right)^2 }{n_x}},\quad 
    \varepsilon_t(x) = \sqrt{\frac{\sum_{j=1}^{n_t} \left(\hat{u}(x,t_j)-u(x,t_j)\right)^2 }{n_t}}.\quad 
\end{equation}
Here, $\hat{u}(x,t)$ denotes an RBF or PINN solution and $u(x,t)$ is the corresponding analytical solution.

For Problems 2 and 3, in which we use experimental measurements of concentrations as a function of time, we evaluate $\varepsilon_t(x)$ at the measurement locations, with $n_t$ equal to the number of measurements.

RBF–RLS solutions are obtained with $N=(N_x+2N_{\text{out}}+1)\times (N_t+2N_{\text{out}}+1)$ RBF centroids centered on a uniform mesh. The number of columns and rows in the grid inside the domain are $N_x$ and $N_t$, respectively, while $N_{\text{out}}$ is the number of rows of RBF centroids outside of the domain. The size of the grid cells is $\Delta_x=1/N_x$ times $\Delta_t=1/N_t$. In all cases, we set the number of quadrature points (residual points or least squares collocation points) to $N_\Omega\approx1.3 N_x$ and $N_T\approx1.3N_t$ (such that $N_{\Omega\times T}^\text{e} = N_\Omega N_T$, $N_{T}^\text{ep}=N_T$, $N_{\partial\Omega\times T}^\text{b}=2N_T$, and $N_{\Omega}^\text{i}=N_{\Omega}$), that provide sufficiently accurate quadrature. 
Compared to much finer quadrature ($N_\Omega \approx 2N_x$ and $N_T \approx 2N_t$), the RMSE increases by less than $1.5\%$, indicating that quadrature errors are negligible relative to other error sources, such as approximation and regression.
In PDE problems with advection, we find that $N_t/N_x\approx v$ yields the best results, especially for high-Peclet-number problems. In the problem without advection, we set $N_t = N_x$. We regularize the RBF–RLS matrix $A$ by adding a small diagonal term $\epsilon I$ (L2 regularization), setting $\epsilon=10^{-6}$ (except for Problem 3 where $\epsilon=10^{-4}$ was needed due to larger matrices). Regularization is important for numerically stabilizing matrix solutions (particularly when using Cholesky decomposition) and for regulating the influence of RBF centroids located outside the domain. In RBF, we always set the squared-residual weights $\lambda^{\text{e}}=\lambda^{\text{b}}=\lambda^{\text{i}}=1$, since we find results are not very sensitive to their variation. Except in the forward solutions in Problem 1, where we compare different values, we set $\frac{h}{\Delta} = \frac{h_x}{\Delta_x}=\frac{h_t}{\Delta_t}=1.4$, which we find in that section to be an appropriate value for the number of centroids $N$ we use in our solutions.

The RBF–RLS solutions of Problems 2 and 3 are compared to the PINN solutions. 
For Problem 1, we were not able to obtain a reasonably accurate PINN solution. For Problem 4, PINN produced a qualitatively correct but substantially less accurate solution than RBF–RLS, so we did not attempt a PINN inverse solution.
Since PINNs are more sensitive to squared-residual weights, we manually tune them in the PINN solutions. For each case, we report the chosen values. In all PINN solutions, we use the Adam optimizer for training. 

We study the effect of measurement noise and model errors on the inverse solutions. In Problem 1, we use ``exact'' synthetic measurements derived from an analytical solution. 
In Problem 4, we use noisy synthetic measurements obtained by multiplying the analytical solution by independent random factors uniformly distributed on [0.9,1.1].
In Problems 2 and 3, we use real-world measurements of concentrations subject to measurement noise and inverse solutions affected by the model error. For Problems 1, 3, and 4, inverse solutions are randomized, with different random realizations corresponding to different initial guesses of the parameters in all three cases, and different randomly generated noise in Problem 4.  We don't randomize the inverse solution for Problem 2 because of how long it takes to obtain PINN solutions.

\subsection{Advection dispersion equation}

\subsubsection{Problem 1: Advection-dispersion equation with a point source initial condition}\label{sec:res_ADE-inst}

Here, we consider the advection-dispersion equation with an instantaneous point release of a tracer with concentration $u$:
\begin{equation}\label{eq:ADE}
\frac{\partial u}{\partial t}+v\frac{\partial u}{\partial x} - D \frac{\partial^2 u}{\partial x^2}=0, \quad x\in(0,1), \quad t\in(0,1]
\end{equation}
subject to the initial condition
\begin{equation}\label{eq:Dirac_IC}
u(x,t=0) = U(x) = M^\text{ip} \delta (x-x^\text{ip})
\end{equation}
and the boundary conditions
\begin{equation}
u(x=0,t)= u(x=1,t)=0,
\end{equation}
where $v$ is the advection velocity and $D$ is the dispersion coefficient. 

By setting $x^{\text{ip}}=0.1$, $v=0.7$, and $D=0.008$, we guarantee that the solution of this PDE problem near the boundaries is close to zero for $t\le1$ and can be accurately approximated by the analytical solution of Eq. \eqref{eq:ADE} defined on an infinite domain, which has the form:
\begin{equation}\label{eq:ADE_Dirac_analyt}
    u(x,t) = \frac{M^\text{ip}}{\sqrt{4\pi D t}} \exp\left(-\frac{(x-x^{\text{ip}}-vt)^2}{4Dt}\right).
\end{equation}

We test the RBF–RLS method for $N_x \in \{30,40,50,60,70,80\}$, $\frac{h}{\Delta}  \in \{1.2,1.4,1.6\}$, and $N_{\text{out}} \in\{0,1,2\}$.

\begin{figure}
    \centering
    \includegraphics[width=1.0\linewidth]{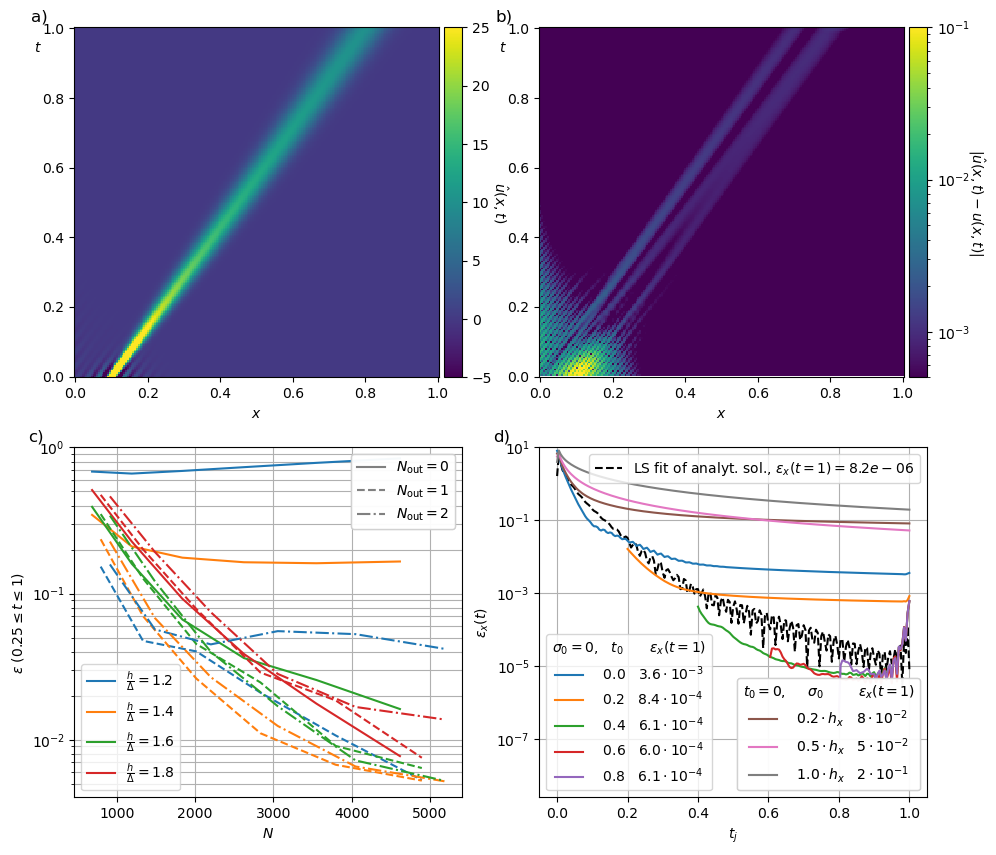}
    \caption{RBF–RLS forward solution of the ADE with an instantaneous point release. (a) Solution obtained using $\frac{h}{\Delta} =1.4$, $N_x = 80$, and $N_{\text{out}}=2$. (b) Point errors of the solution in (a) with respect to the analytical solution. (c) Comparison of global RMSEs (computed for $t\in [0.25,1.0]$) as a function of $N_x$ for different combinations of  $N_{\text{out}}$ and $\frac{h}{\Delta}$. Different colors of RMSE curves correspond to different values of $\frac{h}{\Delta}$. Solid, dashed, and dot-dashed curves correspond to $N_{\text{out}}=0$, 1, and 2, respectively.  (d) Comparison of time-dependent space-averaged RMSEs, $\varepsilon_x(t)$, between the least-squares fit of the analytical solution onto the RBF expansion and various RBF–RLS solutions. Some RBF–RLS solutions are obtained by setting the initial condition at $t_0$ to values ranging from 0 to 0.8, computed on the domain $(t_0,1]$ ($\sigma_0 = 0$). Some are obtained by substituting the Dirac delta initial condition with a Gaussian function of standard deviation $\sigma_0>0$ at $t_0=0$. Errors at the last timestep $\varepsilon_x(t=1)$ are given in the legend for each solution. The blue line corresponds to the RBF–RLS solution with analytical integration of the Dirac delta.   
    }
    \label{fig:ADE_inst_point_rel}
\end{figure}

Figure \ref{fig:ADE_inst_point_rel}a shows $\hat{u}$ as a function of $x$ and $t$ for $\frac{h}{\Delta} =1.4$, $N_x = 80$, and $N_{\text{out}}=2$ (combination that yields the lowest global RMSE). The absolute point errors with respect to the analytical solution are presented in Figure \ref{fig:ADE_inst_point_rel}b. The relative errors (the point errors divided by the maximum concentration at each time) at late times ($t>0.25$ for the selected $\frac{h}{\Delta}$ ratio) are less than 0.001. The relative errors at early times ($t<0.25$) are significantly larger, with a maximum of 0.1 at the first time step near the source. The errors become very small closely after the plume's standard deviation exceeds $h_x$, which, for the selected conditions, occurs at $t=\frac{h_x^2}{2DN_x^2}\approx 0.19$. 

Figure \ref{fig:ADE_inst_point_rel}c compares the global RMSEs $\varepsilon$ for different sets of hyperparameters $N$, $\frac{h}{\Delta}$, and $N_{\text{out}}$. The smallest errors are obtained for $\frac{h}{\Delta}=1.4$, $N_x = 80$, and $N_{\text{out}}=2$, with a total number of RBFs $N=5185$. 

Since errors at early times bias $\varepsilon$, we compute the global RMSEs for $t>0.25$. We observe two trends. First, some centroids outside the domain (i.e., $N_{\text{out}}>0$) are necessary to improve function representation near the boundaries, yet excessive external centroids can increase errors, especially under weak regularization. Second, refining the expansion (increasing $N_x$ and $N_t$, equivalently decreasing $\Delta_x$ and $\Delta_t$) lowers the overall error, but the optimal shape parameter shifts to larger values of $\frac{h}{\Delta}$, indicating that relatively wider, smoother RBFs are needed as the discretization error floor drops. This trade‑off mirrors classical SPH error decompositions—even though our formulation is residual least squares rather than collocation—where the total error behaves asymptotically like the sum of a term that decreases with $h$ (caused by the sharpening of the RBFs) and a term that decreases with $\Delta/h$ (caused by using a finite number of RBFs/particle count)~\cite{Quinlan2006Truncation}. 

In Figure \ref{fig:ADE_inst_point_rel}d, we analyze how the smoothness of the initial condition affects the accuracy of the RBF–RLS solution. To this end, Eq. \eqref{eq:ADE} is solved on the time interval $(t_0,1]$ using the initial condition
\begin{equation}\label{eq:terminal_condition}
u(x,t_0) = \frac{M^\text{ip}}{\sqrt{2\pi (\sigma_0^2+2Dt_0)}} \exp\left(-\frac{(x-x^{\text{ip}}-vt_0)^2}{2(\sigma_0^2+2Dt_0)}\right),
\end{equation}
where $t_0 \in [0,1)$ and $\sigma_0>0$. For $\sigma_0=0$, this corresponds to imposing the analytical ADE solution at time $t_0$. For $t_0=0$, the Dirac delta initial condition is approximated using a Gaussian mollifier with standard deviation $\sigma_0$. The limiting case $t_0=\sigma_0=0$ recovers the Dirac delta initial condition, with an analytical solution given by Eq. \eqref{eq:ADE_Dirac_analyt}. In all cases except the last, the initial condition residual is integrated numerically using $N_\Omega$ uniformly distributed quadrature points.

Figure \ref{fig:ADE_inst_point_rel}d presents the time-dependent RMSE, $\varepsilon_x(t)$, for two families of initial conditions: (i) analytical solutions imposed at times $t_0\in\{0.0,0.2,0.4,0.6,0.8\}$ with $\sigma_0=0$, and (ii) Gaussian mollifiers with widths $\sigma_0\in\{0.2,0.5,1.0\}\cdot h_x$ imposed at $t_0=0$. The figure also includes the least-squares fit of the analytical solution, given by Eq. \eqref{eq:ADE_Dirac_analyt}, onto the RBF expansion. Since the coefficients of this expansion are obtained by minimizing the squared difference between the RBF representation and the analytical solution, the associated global RMSE provides a lower bound for the error attainable by the RBF–RLS solution using the same basis. Nevertheless, the instantaneous error $\varepsilon_x(t)$ of the RBF–RLS solution can locally fall below the least-squares-fit error at certain times.

The results show that both the Dirac delta RBF–RLS solution ($t_0=\sigma_0=0$) and the least-squares-fit RBF approximation exhibit decreasing errors with time as diffusion progressively smooths the analytical solution. At early times, the RBF–RLS errors closely follow the least-squares-fit errors, indicating that the dominant source of error arises from the inability of the RBF basis to accurately represent the sharp initial condition, rather than from the RLS formulation itself. Although diffusion damps most of these errors, some persist (e.g., the diagonal stripes in Figure \ref{fig:ADE_inst_point_rel}b), causing the RBF–RLS errors to deviate from the least-squares-fit errors at later times.

Figure \ref{fig:ADE_inst_point_rel}d also shows that $\varepsilon_x(t)$ decreases as $t_0$ increases up to approximately $t_0\approx0.25$, demonstrating that the RBF–RLS formulation benefits from smoother initial conditions. However, due to errors arising from approximating a Dirac delta with a mollifier, a different behavior is observed when smoothing is introduced through a Gaussian mollifier at $t_0=0$ (the most typical approach for handling Dirac deltas): increasing $\sigma_0$ leads to larger errors. For a fixed number of quadrature points $N_\Omega$, the smallest errors are obtained when the mollifier width satisfies approximately $\sigma_0\sim ||\Omega||/N_\Omega$, as is the case for $\sigma_0=0.5\cdot h_x$ in Figure \ref{fig:ADE_inst_point_rel}d. Increasing $N_\Omega$ would allow smaller values of $\sigma_0$ and, correspondingly, lower errors, approaching the analytically integrated Dirac delta solution asymptotically, albeit at increased computational cost.

Overall, although the RBF representation becomes less accurate for sharper initial conditions, the errors introduced by mollifying the Dirac delta are even larger. Consequently, when the governing physics justify it, directly incorporating the Dirac delta through analytical integration, as proposed in Eqs. \eqref{eq:LLR} and \eqref{eq:LLR-derivatives}, yields the most accurate results.

The PINN method failed to converge for this problem. The PINN results were highly sensitive to the residual weights, and we were unable to simultaneously reduce all residual terms to acceptable levels. Depending on the choice of weights, either the initial and boundary conditions were poorly satisfied, or the interior solution departed substantially from the expected ADE behavior.

Next, we solve the inverse ADE problem for $(v,D)$ assuming that $N^*=n_x=200$ measurements of $u(x,t)$ are available at $t=1$.
The measurements were generated by the analytical solution with the reference values $v=0.7$ and $D=0.0008$. The location of the point source $x^{\text{ip}}=0.1$ and the mass $M^\text{ip}$ are assumed to be known. 
The minimization problem in the RBF–RLS inverse solution \eqref{eq:RBF_inverse} is solved using the \textit{trust-exact} algorithm with analytically computed gradient and Hessian. Initial guesses for $v$ are drawn uniformly within $\pm 8\%$ of the reference value; initial guesses for $D$ are drawn log‑uniformly from $[0.5,\,2]\times$ the true value. Across 30 random initializations, every run converged to the same optimum: $v$ within $0.022\%$ and $D$ within $0.228\%$ of their reference values, with a mean of only 11 forward solves needed to converge. If initial guesses for $v$ were taken outside the stated range, the solution would sometimes converge to unphysical local minima because the plume center is advected beyond the domain.

\subsubsection{Problem 2: Advection-dispersion equation with a step boundary condition}

As a benchmark of the RBF–RLS method against the PINN method, we consider an ADE problem without the delta function in either the source, initial, or boundary conditions. Specifically, we use both methods to solve the ADE \eqref{eq:ADE} on the unitary space--time domain subject to the initial condition:
\begin{equation}
    u(x,t=0) = 0
\end{equation}
and the boundary conditions
\begin{equation}
    u(x=0,t) = 1, \quad \frac{\partial u}{\partial x}\Big|_{x=1} =0.
\end{equation}
Because of the step-function change in $u$ at $(x=0, t=0)$, this PDE problem presents a similar challenge to approximating a solution with large derivatives, but not the difficulties that arise from integrating a Dirac delta term. 

Among other problems, this ADE describes a tracer experiment in a porous medium presented in \cite{levy2003measurement}. 
We solve the forward and inverse ADE problem using RBF–RLS and PINN methods. In the forward problem, we use values for the velocity and dispersion coefficient reported in \cite{levy2003measurement}. 

In the inverse ADE problem, we estimate $v$ and $D$ from measurements of $u(x,t)$ collected over time at the normalized distance $x=1$, i.e., at the right boundary of the spatial domain, as provided in \cite{levy2003measurement}.  

In the forward solution, we normalize the parameters $\bar{v} = 0.213 \frac{\text{cm}}
{\text{min}}$ and $\bar{D} = 0.037\frac{\text{cm}^2}{\text{min}}$ 
estimated in \cite{levy2003measurement} as  $v=\bar{v} \frac{T}{L}$ and $D = \bar{D} \frac{T^2}{L}$, where $L=86 \,\text{cm}$ is the size of the experimental cell in the experiment and $T=805$ min is the time of the last measurement. The time and space coordinates in the ADE model are normalized as $t = \overline{t}/T$ and $x = \overline{x}/L$, where $\overline{t}$ (min) and $\overline{x}$ (cm) are the time and space coordinates in the experiments. 

The forward RBF and PINN solutions are compared to the semi-analytical solution of the ADE problem obtained in the Laplace domain in the form
\begin{equation}
    \tilde{u}=\frac{\alpha_{-} \exp\left(\alpha_{-}+\alpha_{+}x\right)-\alpha_{+} \exp\left(\alpha_{+}+\alpha_{-}x\right)}{s\left(\alpha_{-} \exp\left(\alpha_{-}\right)-\alpha_{+} \exp\left(\alpha_{+}\right)\right)},\quad \alpha_{\pm} =\frac{v\pm\sqrt{v^2+4Ds}}{2D},
\end{equation}
and numerically inverted into the physical domain by the de Hoog algorithm \cite{de_hoog_improved_1982,wang_surface_2021,invlap,reyna_tartakovsky_2025}. 

We choose RBF–RLS and PINN models with a similar number of parameters to ensure a fair comparison between the two methods. In PINN, we use a fully connected network with two hidden layers of 84 neurons each (7,477 trainable parameters).  The PINN is trained with  $N_{\Omega\times T}^\text{e}=2,000$ interior residual points, $N_{\partial\Omega\times T}^\text{b}=1,000$ in the boundaries, and $N_{\Omega}^\text{i} =500$ in the initial condition. The learning rate of the Adam optimizer is initialized at $2.5 \times 10^{-3}$, with a decay of 10\% every 12,000 epochs. To speed up convergence, the same set of random residual points is used throughout training. The network is trained for 1,250,000 epochs. The weights in Eq. 
\eqref{eq:DLLR} for PINN are set to 
 $\lambda^{\text{e}}=\lambda^{\text{b,d}}=\lambda^{\text{i}}=1$ and $\lambda^{\text{b,u}}=10$, where $\lambda^{\text{b,d}}$ and $\lambda^{\text{b,u}}$ correspond to the right and left boundary squared residuals respectively.

The RBF–RLS solution uses $N_x=55$, $N_t=109$ and $N_{\text{out}}=4$ (7,552 RBF coefficients).

\begin{figure}
    \centering
    \includegraphics[width=1.0\linewidth]{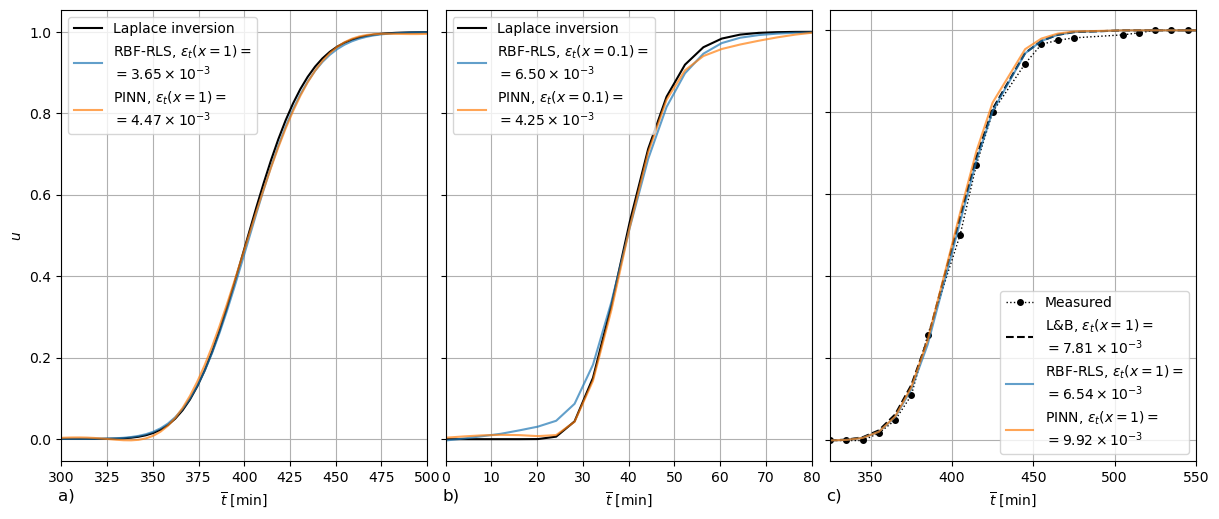}
    \caption{Comparison of RBF–RLS and PINN forward and inverse solutions of the ADE with a step boundary condition. a) Forward solutions at $\bar{x}=L=86\,\text{cm}$ b) Forward solutions at $\bar{x}=L/10=8.6\,\text{cm}$ c) Laplace solution with parameters estimated by Levy and Berkowitz (L\&B) \cite{levy2003measurement}, RBF–RLS, and PINN, at $\bar{x}=L=86\,\text{cm}$.}
    \label{fig:Berkowitz}
\end{figure}

Figure \ref{fig:Berkowitz} compares the forward and inverse solutions obtained using PINN and RBF–RLS. The forward RBF–RLS and PINN solutions have similar global RMSEs (computed for the whole space--time domain) of $4.73\times10^{-3}$ and $4.94\times10^{-3}$, respectively. However, the space-dependent time-averaged RMSEs $\varepsilon_t(x)$ (computed at the times of measurement) are significantly different in the two methods. At $x=0.1$, the RBF–RLS $\varepsilon_t$ error is approximately 50\% higher than that of PINN, whereas at $x=1$ it is about 20\% lower (Figures \ref{fig:Berkowitz}b and \ref{fig:Berkowitz}a, respectively). The RBF–RLS achieves higher accuracy at $x=1$, which is important for solving the inverse problem because $ x=1$ is the location of the measurements. 

Figure \ref{fig:Berkowitz}c compares reconstructed solutions of $u(\overline{x}=L,\overline{t})$ obtained with the semi-analytical solution using parameters $(v,D)$ estimated by RBF–RLS and PINN methods and estimated in \cite{levy2003measurement}, as well as the measured values of $u$. The RMSE $\varepsilon_t$ between the measured and predicted $u$ is the lowest in RBF–RLS, with PINN yielding 51\% larger error and the parameters in  \cite{levy2003measurement} producing  19\% larger RMSE. The RMSE values in all methods are reported in Figure \ref{fig:Berkowitz}c. 
Table \ref{tab:estimated_parameter} lists $\overline{v}$ and $\overline{D}$ parameters
estimated from the RBF–RLS and PINN inverse methods, as well as the parameters estimated in \cite{levy2003measurement}.

\begin{table}
    \caption{Parameters $\overline{v}$ and $\overline{D}$,
estimated from the RBF–RLS and PINN inverse methods as well as the parameters reported in \cite{levy2003measurement}}
    \centering
    \begin{tabular}{ccc}
         & $\overline{v}$, $\frac{\text{cm}}{\text{min}}$ & $\overline{D}$, $  \frac{\text{cm}^2}{\text{min}}$ \\
      RBF–RLS  & 0.2126 & 0.0358\\
      PINN   & 0.2134 & 0.0341\\
      Levy and Berkowitz \cite{levy2003measurement} & 0.213 & 0.037 \\
    \end{tabular}
    \label{tab:estimated_parameter}
\end{table}

In the inverse PINN solution, we use the same setup as in the forward problem and set $\lambda_{*}=1,000$. The unknown parameters $v$ and $D$ are fixed to their initial guess for the first 30,000 epochs and become variable thereafter, once a physically plausible solution has been reached. When this wasn't done, the model parameters were overfit to unphysical solutions. This is not necessary in RBF-RLS, where each loss during model parameter optimization corresponds to an optimal RBF-RLS forward solution. The initial guess for the parameters was set to $0.8\bar{v}^*$  and $1.5 \bar{D}^*$, where  $\bar{v}^*$ and $\bar{D}^*$ are the values estimated in \cite{levy2003measurement}. 
The RBF–RLS solutions show stronger dependence on the initial parameter guess than the PINN inverse solutions. On the other hand, PINN inverse solutions are sensitive to the value of $\lambda_{*}$ (which doesn't exist in RBF–RLS), with smaller $\lambda_{*}$ yielding larger $\bar{D}$. 

\subsection{Problem 3: Advection-dispersion equation with first-order mobile-immobile exchange}\label{sec:prob_ADE-exch}

In this section, we consider the advection-dispersion equation with first-order mobile-immobile exchange
\begin{align}
    &\frac{\partial u_{\text{m}}}{\partial t} +v \frac{\partial u_{\text{m}}}{\partial x} - D\frac{\partial^2 u_{\text{m}}}{\partial x^2} = - \beta k \left( u_{\text{m}} - u_{\text{im}}\right),
    \\\nonumber
   & u_{\text{m}}(t=0) = M^{\text{ip}}\, \delta(x),\quad u_{\text{m}}(x=-\infty) = u_{\text{m}}(x=\infty) = 0,
    \\\nonumber
   & \frac{\partial u_{\text{im}}}{\partial t}  = k \left(u_{\text{m}} -  u_{\text{im}}\right), \quad u_{\text{im}}(t= 0) = 0,
\end{align}
where $u_{\text{m}}$ is the concentration in the mobile phase, $u_{\text{im}}$ is the concentration in the immobile phase, $v$ is the effective advection velocity, $D$ is the effective dispersion coefficient, $\beta$ is the effective area ratio between the immobile and mobile phases, $k$ is the exchange rate between phases, and $M^{\text{ip}}$ is the injected mass per unit area (released at location $x=0$). Such equations have been used to describe transport of solutes in rivers and porous media \cite{seo_moment-based_2001,gonzalez-pinzon_scaling_2013,wang_surface_2021,aghababaei_temporal_2023,reyna_tartakovsky_2025}.
 
This system of equations has a solution in the Laplace domain \cite{reyna_tartakovsky_2025}
\begin{equation}
    \tilde{u}_{\text{m}}=\frac{M^{\text{ip}}}{\sqrt{v^2+4Ds\left(1+\frac{\beta k}{s+k}\right)}} \exp\left(\frac{v (x-x^*) }{2D} -\frac{\sqrt{v^2+4Ds\left(1+\frac{\beta k}{s+k}\right)}|x-x^*|}{2D} \right),
\end{equation}
where $\tilde{u}_{\text{m}}$ is the Laplace transform of the mobile phase concentration. We numerically invert the Laplace domain solution using the method by de Hoog \cite{de_hoog_improved_1982,wang_surface_2021,invlap,reyna_tartakovsky_2025}.

We can write the PDE operator as
\begin{equation}
    \mathcal{L} =\begin{bmatrix}
       \frac{\partial }{\partial t} + v \frac{\partial }{\partial x} - D \frac{\partial^2 }{\partial x} +  \beta k& - \beta k\\
        - k & \frac{\partial }{\partial t} +  k
    \end{bmatrix} ,
\end{equation}
where $\mathcal{L}$ is linear to $\mathbf{v}=[M_0, v,  D, \beta k, k]$. The vector of coefficients $\mathbf{u}$ is formed by stacking the vectors of coefficients of each of the two state variables (mobile and immobile). Using this definition of the operator, the squared residuals of each of the two equations in the system are added together with equal weights.

For the inverse problem, we fit the model to the first breakthrough curve measured in Antietam Creek on 5/27/1969 by the USGS \cite{nordin_empirical_1974,rodriguez_tierras_2025}. Measurements are taken at a single location over time in the mobile zone (river channel). 

For this problem, we set $x^{\text{ip}}=0.1$ and scale time and space such that the sampling location is fixed at $x^*=0.45$ and its measured peak occurs at $t=0.5$. This ensures consistency with the solution defined on an infinite domain and captures a significant portion of the tail within the domain.

We set $N_x=50$, $N_t=35$, and $N_{\text{out}}=2$. For parameter estimation, we compare three optimization algorithms \textit{Nelder-Mead} (no gradient or Hessian), \textit{L-BFGS-B} (analytical gradient), and \textit{trust-exact} (analytical gradient and Hessian). To ensure comparable computational cost, we limit the number of forward evaluations to 150 for \textit{Nelder-Mead}, and the number of iterations to 38 and 20 for \textit{L-BFGS-B} and \textit{trust-exact}, respectively, resulting in an average runtime of approximately 250 seconds per optimization. 

Initial parameter guesses (in domain-scaled units) are set to $v=0.7$, $D=0.0008$, $\beta=0.05$, and $k=10.0$. The initial value of $M_0$ is chosen such that the integral of the breakthrough curve generated from the initial parameters matches that of the measured data. To assess consistency of convergence, initial parameter guesses for $(M_0,v,D,\beta,k)$ are perturbed in 5 random realizations, the first three parameters log-uniformly between plus or minus $10\%$, $10\%$, and $30\%$, respectively, and the last two log-uniformly within a factor of 2. 

\begin{figure}
    \centering
    \includegraphics[width=1.0\linewidth]{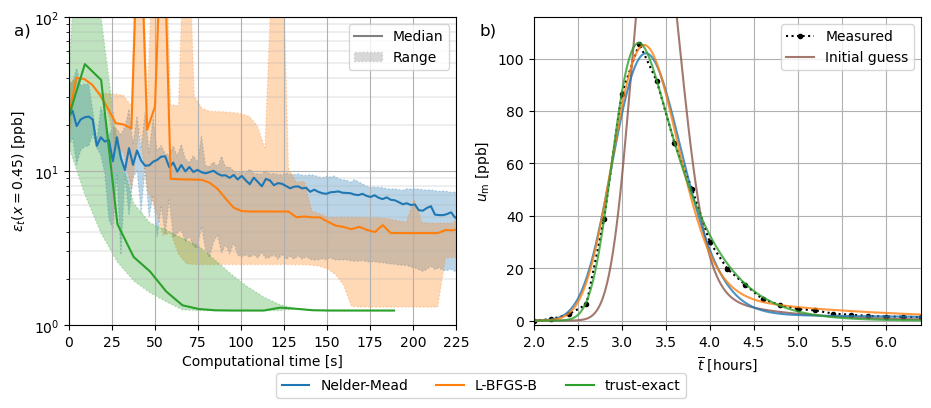}
    \caption{RBF–RLS inverse solution of the ADE with first order exchange to an immobile zone fitted to a breakthrough curve measured in Antietam Creek on 5/27/69 \cite{nordin_empirical_1974,rodriguez_tierras_2025}. a) Evolution of the RMSE of the inverse solution (RBF–RLS forward solution with estimated parameters) during optimization of the parameters. Median and range (minimum to maximum) for 5 random initializations. b) Reconstructed breakthrough curves for the median RMSE classes each.
    }
    \label{fig:river_transport}
\end{figure}

Figure \ref{fig:river_transport}a shows the evolution of the RMSE ($\varepsilon_t(x=x^*)$ computed at the times of measurement) of the inverse solution during the optimization process defined in Eq. \eqref{eq:RBF_inverse}.
The \textit{trust-exact} method achieves the best performance, converging rapidly to a unique solution. In contrast, \textit{Nelder–Mead} and \textit{L-BFGS-B} converge more inconsistently, taking longer and exhibiting greater variability across runs. The \textit{trust-exact} method gives an error in the inverse solution of $\varepsilon_{t}(x=x^*=0.45)=1.23$ ppb (parts per billion) and an error of the reconstructed solution (semi-analytical solution using estimated parameters) of $1.28$ ppb. This indicates that the corresponding forward error (calculated between the inverse solution and the reconstructed solution) of $0.24$ ppb has a minor impact on the inverse solution.

Table \ref{tab:river-transport} shows the parameters of the median error solutions (inverse solution with the median error across different random initial guesses of the parameters) for the three different optimization methods. 
Together with the corresponding reconstructed breakthrough curves in Figure \ref{fig:river_transport}b, Table \ref{tab:river-transport} shows that all three optimization methods reproduce the location and shape of the main breakthrough peak, but they differ substantially in their estimates of the exchange parameters $\beta$ and $k$, which primarily affect the late-time tail. The trust-exact method gives the lowest median RMSE and the most consistent convergence across random initializations.
\begin{table}
    \centering
    \begin{tabular}{|c|c|c|c|c|c|}\hline
         &  $M_0$&  $v$&  $D$&  $\beta$& $k$\\\hline
         Initial guess&  92.21106 & 0.75958 & 0.00620 & 0.03449 & 1.27632 \\\hline
         Nelder-Mead&  88.74203 & 0.77989 & 0.01459 & 0.03770 & 0.20740 \\\hline
         L-BFGS-B&  84.88548 & 0.78425 & 0.01293 & 0.05049 & 0.59751\\\hline
         trust-exact&  90.47050 & 0.84049 & 0.00619 & 0.48559 & 3.91106\\ \hline
    \end{tabular}
    \caption{Parameter estimates (in domain-scaled units) of the ADE with first order exchange to an immobile zone fitted to a breakthrough curve measured in Antietam Creek on 5/27/69 \cite{nordin_empirical_1974,rodriguez_tierras_2025}. Values correspond to the median RMSE solution.}
    \label{tab:river-transport}
\end{table}

When we tested different optimization methods for the other inverse problems (results not shown in this paper), the relative rates of convergence were similar, with the \textit{trust-exact} algorithm converging about twice as fast as the other algorithms. Relative times were consistent despite the different number of dimensions. However, for lower-parameter problems, we observed low variability across runs for all methods.

As was the case for the ADE with a point source, and for similar reasons, we couldn't obtain reasonable solutions using PINNs.

\subsection{Problem 4: Diffusion equation with point source forcing} \label{sec:prob_gwflow}

Here, we consider a diffusion equation (DE) with point source forcing. We set  $\mathcal{L}(u)=\frac{\partial u}{\partial t} - k \frac{\partial^2 u}{\partial x^2}$, $f= Q\,\delta(x-x^\text{ep})$, $\mathcal{B}_l(u)=u$ and $g_l=1$ at $x=0$ and $x=1$, and $U=1$. This problem could represent, among other systems, the evolution of the hydraulic head $u$ in a one-dimensional confined aquifer under a constant pumping rate $Q$. We set $k=0.1$ and $Q=1$.
An analytical solution to this problem is given in \ref{app:analytical_PS}. We compare the PINN and RBF–RLS forward solutions. Next, we obtain the RBF–RLS inverse solution using noisy synthetic data. Due to PINN's failure in obtaining an accurate forward solution, we don't attempt to obtain a PINN inverse solution.  

The RBF–RLS forward and inverse solutions use $N_x=80$, $N_t=80$ and $N_{\text{out}}=3$ (7,056 RBF coefficients).
In the PINN solution, we use a fully connected network with 2 layers and 128 neurons in each layer (17,025 learned parameters). Residual weights are set to $\lambda^{\text{e}}=0.1$ and $\lambda^{\text{b}}=\lambda^{\text{i}}=1$. There are $N_{\Omega\times T}^\text{e}=40,000$ interior residual points, $N_{T}^\text{ep}=10,000$ for the Dirac delta residual, $N_{\partial\Omega\times T}^\text{b}=10,000$ in the boundary, and $N_{\Omega}^\text{i}=10,000$ in the initial condition. The learning rate of the Adam optimizer is set to $ 10^{-4}$. New random sets of collocation points are used in each epoch to avoid overfitting, which would otherwise be very severe. We train for a total of 20,000 epochs.

\begin{figure}
    \centering
    \includegraphics[width=1.0\linewidth]{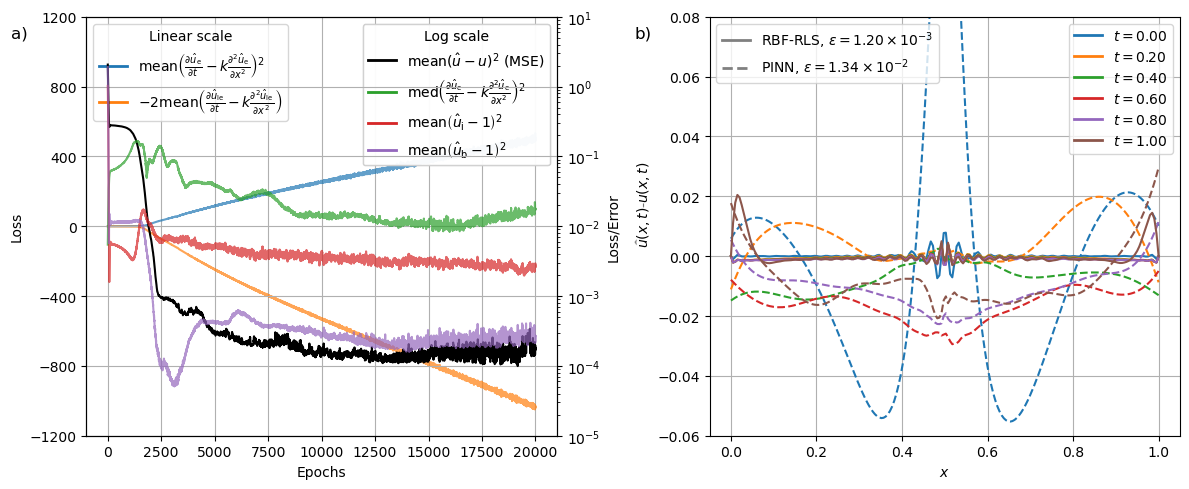}
    \caption{PINN results for the DE with point source forcing. a) Evolution of the losses in PINN. Mean PDE loss and Dirac delta term loss are shown in a linear scale due to their linear asymptotic behavior. The rest of the losses, including the median PDE loss (i.e., the median value of the PDE loss across all residual points) are shown with vertical log scale due to their exponential decay behavior.  
    b) Comparison of RBF–RLS and PINN point errors.}
    \label{fig:PINN_point_source_forcing}
\end{figure}

Figure \ref{fig:PINN_point_source_forcing}a shows the evolution of the losses and errors in the PINN solution, and Figure \ref{fig:PINN_point_source_forcing}b compares the point errors to those of RBF–RLS.  Despite using fewer than half as many learned parameters, RBF–RLS achieves errors that are an order of magnitude smaller. 

This discrepancy is unlikely to stem from limited neural network representational capacity, given the PINN's large parameter count, nor from overfitting, given the number and randomization of quadrature points. Instead, the degradation in PINN's performance appears to stem from the training dynamics of the Dirac delta term. As shown in Figure \ref{fig:PINN_point_source_forcing}a, the squared residuals for the boundary and initial conditions, as well as the median of the PDE loss (which, unlike the mean, excludes effects of points close to the Dirac delta), decay rapidly at first but soon become stagnant. Meanwhile, the Dirac delta contribution is initially constant but then starts decreasing linearly. This behavior is explained in detail in Section \ref{sec:PINN-failure}.

In the inverse RBF solution, we estimate $k$ from noisy measurements of $u$, which are generated by sampling the analytical solution at $N^*=51$ uniformly spaced locations at $t=1.0$ and multiplying them by random values sampled from the uniform distribution on the interval $[0.9, 1.1]$. The inverse problem is solved $30$ times with different random realizations of synthetic data and initial guesses of $k$, log-uniformly distributed in $(0.05,0.2)$. This interval was selected to span values ranging from one-half to twice the true parameter value, providing a broad range of initial guesses to properly assess convergence. Randomization of both the synthetic measurements and the initial guesses is done to evaluate if the inverse solution consistently converges to the correct parameter value. We solve the inverse minimization problem, Eq. \eqref{eq:RBF_inverse}, using the \textit{trust-exact} algorithm (with analytical gradient and Hessian). We represent the results as percentiles across the 30 random realizations (of noisy data and initial guesses) written as (10\%, 50\%, 90\%). 

Table \ref{tab:DE_df_inverse_results} reports pairwise Root Mean Square Differences (RMSD) between the ground truth $u$, the noisy observations, the inverse solution, and the model predictions obtained from the analytical solution evaluated with both the estimated and initial values of $k$. These comparisons enable interpretation of the total inverse error relative to the noise level and allow its decomposition into contributions from parameter estimation and forward-model approximation errors.
allow the total error of the inverse solution to be interpreted relative to the level of noise and decomposed into contributions from parameter estimation and forward-model approximation. 
 RMSDs are defined analogous to $\varepsilon_{x}(t=1)$ as
\begin{equation}
   RMSD (u_a,u_b)= \sqrt{\frac{\sum_{i=1}^{n_x} \left(u_a(x_i,t=1)-u_b(x_i,t=1)\right)^2 }{n_x}}, 
\end{equation}
where $(u_a,u_b)$ are any pair of solutions or data. Values of the first column of Table \ref{tab:DE_df_inverse_results}, which are obtained by comparing to the noiseless ground truth, reduce to RMSEs.

RMSDs involving the analytical reconstructions isolate errors due to incorrect parameter values (including the initial guess), while differences between the inverse solution and its analytical counterpart quantify the additional error introduced by the RBF–RLS forward solver.

From the values of Table \ref{tab:DE_df_inverse_results}, we see that the estimated parameters in the analytical solution decrease the error with respect to the noiseless ground truth by approximately an order of magnitude relative to the initial parameter guesses. These errors are also half as large as the noise, whose magnitude, in terms of RMSDs, is given by the first cell of the table, indicating it has been successfully filtered. The RMSD between the inverse solution and the noisy data is approximately equal to the magnitude of the noise, suggesting that the inverse solution is approaching the accuracy limit imposed by the noise level (since the inverse residual of Eq. \eqref{eq:RBF_inverse} could not be further reduced).

From comparing the last two rows in Table \ref{tab:DE_df_inverse_results}, we see that the difference between $\hat{u}$ obtained from the inverse solution and the analytical solution with the estimated $k$ is negligible in comparison to other sources of error. This suggests that the RBF–RLS forward solver is sufficiently accurate for the purposes of this inverse problem.

The relative error in the estimation of $k$ reduces to (0.633\%, 3.664\%, 9.584\%), compared to (6.298\%, 40.049\%, 80.883\%) for the initial guesses. The former is uncorrelated with the latter (Pearson correlation coefficient of $-0.03$), indicating that convergence is consistent across the range of initial guesses.

\begin{table}
    \centering
    \begin{tabular}{|>{\raggedright\arraybackslash}p{0.21\linewidth}|>{\centering\arraybackslash}p{0.16\linewidth}|>{\centering\arraybackslash}p{0.16\linewidth}|>{\centering\arraybackslash}p{0.16\linewidth}|>{\centering\arraybackslash}p{0.16\linewidth}|}\hline
      &Ground truth (noiseless data)& Noisy data&  Analytical recons. w/ init. guess $k$&Inverse solution\\\hline
 Noisy data& (0.098, 0.108, 0.116)& ---&  ---&---\\\hline
 Analytical recons. w/ init. guess $k$& (0.025, 0.162, 0.260)& (0.102, 0.193, 0.277)& ---&---\\\hline
 Inverse solution& (0.005, 0.018, 0.040)& (0.098, 0.106, 0.114)&  (0.018, 0.158, 0.259)&---\\\hline
 Analytical recons. w/ est. $k$& (0.003, 0.015, 0.037)& (0.096, 0.106, 0.114)&  (0.018, 0.159, 0.258)&(0.004, 0.004, 0.008)\\ \hline
    \end{tabular}
    \caption{Root Mean Square Differences (analogous to $\varepsilon_x(t=1)$) between pairs of solutions of the DE with point source forcing. In each cell, the 10th, 50th, and 90th percentiles (over the realizations of random noise and initial guess) of the RMSDs are shown in parentheses. The complete RMSD matrix is symmetric with zeros on the diagonal. Only non-zero and non-repeated values are shown here. Noiseless data represents the ground truth, obtained using the analytical solution with $k=0.1$. Noisy data is obtained by adding noise to the ground truth. We also compare the analytical reconstruction of the solution using the initial guess of $k$ and the estimated value of $k$. The inverse solution is the RBF–RLS forward solution obtained using estimated parameters. 
    }
    \label{tab:DE_df_inverse_results}
\end{table}

We perform the detailed RMSD decomposition only for this example because it combines noisy synthetic data with a known analytical ground truth and a single estimated parameter. This allows the total inverse error to be separated into measurement noise, parameter-estimation error, and RBF–RLS forward-solver error. Such a decomposition is less informative for the problem reported in Section \ref{sec:prob_ADE-exch}, where the ground truth solution and true parameters are unknown, and was omitted for the ADE point-source example because its inverse solution was obtained using noiseless data.

\subsubsection{Why PINN fails for PDEs with point sources }\label{sec:PINN-failure}

PINNs struggle with PDEs containing Dirac delta sources because the Neural Tangent Kernel (NTK) induces global coupling of residuals, preventing simultaneous satisfaction of the singular forcing and the smooth PDE constraints. While the forward PINN solution for the DE is qualitatively correct, its errors are much larger than those of RBF–RLS, despite PINNs having twice as many trainable parameters. This formulation of PINNs was particularly sensitive to the training weights, and solutions could only be obtained with a small learning rate ($10^{-4}$). No stable or accurate PINN solutions could be obtained for the other problems with Dirac deltas considered in this paper. 

We explore the failure of PINN solutions for problems with Dirac delta sources using NTK theory. In this context, failure refers to stagnation in PDE, boundary, and initial-condition losses, preventing any further reduction in errors despite continued training or increased complexity of the Neural Networks. 

Unlike standard PINN formulations where all residuals appear squared, the Dirac term introduces a linear (non-squared) contribution to the loss (Eq. \eqref{eq:LLR}). The training process (via gradient flow in parameter space) can be analyzed using NTK theory, resulting in a dynamical system \cite{Wang2022NTK}. We derive the form of the dynamical system for PDE problems with Dirac delta forcing in \ref{app:NTK}. In the infinite-width-network limit, NTK converges to a constant-kernel limit, making the dynamical system linear. A limiting NTK is expected to exist in this setting, as the definition of the kernel, the types of residuals, and the neural network architecture do not qualitatively differ from those analyzed in Wang et al. \cite{Wang2022NTK}. The functional form of the NTK depends on the exact form of the losses, the architecture, the activation function, and the initialization.

In \ref{app:NTK}, we show that, in the dynamical system representing training, the residual associated with the Dirac delta source does not converge to zero. Instead, it decreases linearly in training time, while all other residuals converge to nonzero constants. This is in contrast to exponential decay toward zero expected in typical PINN formulations \cite{Wang2022NTK}. This behavior agrees with the trends in the losses observed in Figure \ref{fig:PINN_point_source_forcing}, which shows, after 2500 epochs, a linear decrease of the Dirac delta loss and stagnation of the rest of the losses. The particular dynamics are a consequence of forcing terms appearing in the dynamical system, caused by interactions between the Dirac delta term residual and the rest of the residuals. 

These interactions can be interpreted directly in terms of the NTK structure. For example, the two non-constant terms resulting from the squared residual of the PDE in Eq. \eqref{eq:LLR} are $R^{\text{e}}=\mathcal{L}(\hat{u})^2 $ and $R^{\text{ep}}=-2  M^{\text{ep}}(t)\mathcal{L}(\hat{u}(\mathbf{x}^{\text{ep}}))$. We define a residual $r=\mathcal{L}(\hat{u})$, and let $K(\mathbf{x}_i,t_i,\mathbf{x}_j,t_j)$ denote the NTK. The kernel quantifies how residuals at two locations $(\mathbf{x}_i,t_i)$ and $(\mathbf{x}_j,t_j)$ co-vary under infinitesimal changes in the network parameters, i.e., how variations driven by $r(\mathbf{x}_i,t_i)$ affect the residual at $r(\mathbf{x}_j,t_j)$.  Since $R^{\text{e}} = r^2$ and $R^{\text{ep}}\propto r$, if $\hat{u}$ matches the true solution, we should see the following distributional structure: $r(\mathbf{x}=\mathbf{x}^{\text{ep}})=\pm\infty$ and  $r(\mathbf{x}\neq\mathbf{x}^{\text{ep}})=0$. However, in the infinitely-wide-network limit the NTK reaches a limit function with nonzero radius of influence (i.e., the distance between a pair of points at which the value of the Kernel is close to zero) \cite{jacot2018neural} so that residuals $r(\mathbf{x}=\mathbf{x}^{\text{ep}})$ are globally coupled to $r(\mathbf{x}\neq\mathbf{x}^{\text{ep}})$. Thus, under the classical PINN architecture and the NTK-limiting assumptions considered here, increasing network width (which, according to the universal approximation theorem, allows representing the solution arbitrarily closely) does not remove this coupling. Therefore, the residual field cannot approach the desired distributional structure simply by increasing the number of network parameters, leading to PINN failure. In terms of the dynamical system, the forcing terms, which represent the limit of the residuals in training, cannot be made arbitrarily close to zero.

In contrast, kernels in RBF–RLS (for compact support RBFs or RBFs decaying to zero at infinity) have a radius of influence proportional to $h$. Since we can make $h\to 0$ as the number of terms in the expansion increases (while also making $\frac{\Delta}{h}\to 0$), forcing terms in the dynamical system can be made arbitrarily small, though numerical limitations exist. Consequently, errors in RBF–RLS solutions can be reduced by increasing the number of trainable parameters.

We note that this limitation of PINNs is derived under the classical PINN architecture, often called vanilla PINN \cite{baty2023solvingstiffordinarydifferential}. More expressive PINN variants, such as those incorporating Fourier feature embeddings, can alter the effective kernel and improve the representation of localized features. These approaches may alleviate the difficulty of representing point-source forcing, although the extent to which they would improve solutions to problems including Dirac delta sources remains an open question.

\section{Conclusions}\label{sec:conclusions}

In this work, we formulated the RBF–RLS method for forward and inverse PDEs with Dirac delta sources. By explicitly integrating singular sources at the residual level, we demonstrated that such problems can be solved without approximating Dirac deltas with Gaussian functions, avoiding a common and often significant source of modeling error in PINNs and other methods. 

For linear PDEs, the RBF–RLS method provides a deterministic linear system whose solution directly minimizes the residual norm. Across all test cases considered, RBF–RLS exhibited low errors, limited sensitivity to hyperparameters, and reliable inverse parameter estimates, including when applied to real-world measurements. The availability of analytical gradients and Hessians improved computational efficiency in inverse problems.

The closely related PINN method, which uses DNNs instead of RBFs to approximate the state variables, produced significantly less accurate solutions for the DE and did not converge for the remaining problems that included Dirac delta sources. Neural Tangent Kernel analysis of PINN for PDEs with point sources revealed that while loss terms from Dirac-delta sources declined linearly, other terms remained stagnant. This analysis also shows that increasing the complexity of DNNs is unlikely to yield further improvements beyond a certain point.

For the ADE problem without a Dirac delta, we find that the RBF–RLS and PINN models with the same number of trainable parameters yield similarly accurate forward solutions, but the RBF–RLS model provides a more accurate estimate of the parameters in the inverse solution. 

While PINNs remain attractive for high-dimensional or strongly nonlinear problems due to their representational flexibility and favorable scaling with dimensionality, our results indicate that RBF–RLS provides a reliable and computationally efficient alternative for diffusion and advection-diffusion equations with point or instantaneous sources. Forward and inverse problems involving sharp fronts, early-time behavior, or high-Peclet-number flows particularly benefit from the linear structure and stability of the RBF–RLS approach.

\bibliographystyle{elsarticle-num} 
\bibliography{RBF_RLS_PINN_delta}
\include{appendices}

\end{document}

%% file: appendices.tex
\appendix

\section{General formulation of Dirac delta source terms}\label{app:general_dirac_delta} 

We can include point or instantaneous sources by adding Dirac delta terms to the right hand side of the PDE or the boundary or initial conditions (Eq. \eqref{eq:general PDE}),  $f=\hat{f}(\mathbf{x},t) + \sum_\alpha M^{\text{ep},\alpha}(t) \delta(\mathbf{x}-\mathbf{x}^{\text{ep},\alpha})+\sum_\alpha M^{\text{ei},\alpha}(\mathbf{x}) \delta(t-t^{\text{ei},\alpha})+\sum_\alpha M^{\text{epi},\alpha} \delta(\mathbf{x}-\mathbf{x}^{\text{epi},\alpha})\delta(t-t^{\text{epi},\alpha})$, $g=\hat{g}(\mathbf{x},t) + \sum_\alpha M^{\text{bi},\alpha}(t) \delta(\mathbf{x}-\mathbf{x}^{\text{bi},\alpha})+\sum_\alpha M^{\text{bi},\alpha}(\mathbf{x}) \delta(t-t^{\text{bi},\alpha})+\sum_\alpha M^{\text{bpi},\alpha} \delta(\mathbf{x}-\mathbf{x}^{\text{bpi},\alpha})\delta(t-t^{\text{bpi},\alpha})$, and $h(\mathbf{x}) = \hat{h}(\mathbf{x}) +  \sum_\alpha M^{\text{ip},\alpha} \delta(\mathbf{x}-\mathbf{x}^{\text{ip},\alpha})$. Here, ``ep'' is a point source in the PDE, ``ei'' an instantaneous source in the PDE, ``epi'' an instantaneous point source in the PDE, ``bp'' is a point source in the boundary, ``bi'' an instantaneous source in the boundary, ``bpi'' an instantaneous point source in the boundary, and ``ip'' a point source in the initial condition. 

The PDE, boundary condition, and initial condition squared loss terms, from Eq. \eqref{eq:LLR-complete}, are, respectively,
\begin{align}\label{eq:integrated_deltas}
    \mathcal{R}^{\text{e}} =& \frac{1}{||\Omega|| T }  \left[\int_\Omega\int_0^T  [\mathcal{L}(\hat{u})^2 - 2\mathcal{L}(\hat{u}) \hat{f}] dt d\mathbf{x}-2 \sum_{\alpha}\int_0^T  M^{\text{ep},\alpha}(t)\mathcal{L}(\hat{u}(\mathbf{x}^{\text{ep},\alpha})) dt \right.
    \\\nonumber
    &\left.\quad\quad\quad\quad-2 \sum_{\alpha}\int_{\Omega} M^{\text{ei},\alpha}(\mathbf{x})\mathcal{L}(\hat{u}(t^{\text{ei},\alpha})) d\mathbf{x} -2\sum_\alpha M^{\text{epi},\alpha}\mathcal{L}(\hat{u}(\mathbf{x}^{\text{epi},\alpha},t^{\text{epi},\alpha})) \right.
    \\\nonumber
   &\left.\quad\quad\quad\quad+\hat{\Delta}^{\text{e}}+ \Delta^{\text{e}} \right],
    \\ \nonumber
    \mathcal{R}^{\text{b}} =& \frac{1}{||\partial\Omega|| T }  \left[\int_{\partial\Omega}\int_0^T  [\mathcal{B}(\hat{u})^2- 2\mathcal{B}(\hat{u}) \hat{g}] dt d\mathbf{x}-2 \sum_{\alpha}\int_0^T  M^{\text{bi},\alpha}(t)\mathcal{B}(\hat{u}(\mathbf{x}^{\text{bi},\alpha})) dt \right.
    \\\nonumber
    &\left.\quad\quad\quad\quad-2 \sum_{\alpha}\int_{\partial\Omega} M^{\text{bi},\alpha}(\mathbf{x})\mathcal{B}(\hat{u}(t^{\text{bi},\alpha})) d\mathbf{x} -2\sum_\alpha M^{\text{bpi},\alpha}\mathcal{B}(\hat{u}(\mathbf{x}^{\text{bpi},\alpha},t^{\text{bpi},\alpha})) \right.
    \\\nonumber
    &\left.\quad\quad\quad\quad+\hat{\Delta}^{\text{b}} + \Delta^{\text{b}} \right],
    \\ \nonumber
    \mathcal{R}^{\text{i}} =&  \frac{1}{||\Omega|| } \left[\int_\Omega  [\hat{u}(t=t_0)^2-2\hat{u}(t=t_0)\hat{h}] d\mathbf{x} -2 \sum_\alpha M^{\text{ip},\alpha} \hat{u}(\mathbf{x}^{\text{ip},\alpha},t_0) +\hat{\Delta}^{\text{i}} + \Delta^{\text{i}}\right].
\end{align}
Here, $\hat{\Delta}^{\text{e}}$, $\hat{\Delta}^{\text{b}}$, and $\hat{\Delta}^{\text{i}}$ are the terms that include $\hat{f}$, $\hat{g}$, or $\hat{h}$, respectively, but not $\mathcal{L}(\hat{u})$, $\mathcal{B}(\hat{u})$, or $\hat{u}(t=t_0)$; therefore, they are independent of $\boldsymbol{\theta}$ and can be droped when solving the minimization problem in Eq \eqref{eq:LLR}. The terms $\Delta^{\text{e}}$, $\Delta^{\text{b}}$, and $\Delta^{\text{i}}$ are also independent of $\boldsymbol{\theta}$ but include integrals of the type $\int_a^b \delta(x)^2 dx$, that are undetermined. These integrals can be approximated by treating the Dirac delta function as a distribution $\delta(x)=\lim_{\epsilon \to 0} \frac{\exp(-x^2/(2\epsilon^2))}{\sqrt{2\pi \epsilon^2}}$. 
Then, for any nonzero but arbitrarily small value of $\epsilon$,   $\Delta^{\text{e}}$, $\Delta^{\text{b}}$, and $\Delta^{\text{i}}$ are finite, and can also be ignored in solving the minimization problem.

Finally, the solution of the minimization problem in Eq \eqref{eq:LLR} with point sources satisfies the zero gradient condition
\begin{align} \label{eq:linear_system}
    \mathbf{0} =& 
    \frac{2\lambda^{\text{e}}}{||\Omega|| T } \left[ \int_\Omega \int_0^T  [\mathcal{L}(\hat{u})-\hat{f}(\mathbf{x},t) ]\frac{\partial \mathcal{L}(\hat{u})}{\partial \boldsymbol{\boldsymbol{\theta}}} dt d\mathbf{x} - \sum_\alpha \int_0^T  M^{\text{ep},\alpha}(t)  \frac{\partial \mathcal{L}(\hat{u}(\mathbf{x}^{\text{ep},\alpha}))}{\partial \boldsymbol{\boldsymbol{\theta}}} dt \right.
    \\\nonumber
    &\left.\qquad\qquad- \sum_\alpha \int_{\Omega} M^{\text{ei},\alpha}(\mathbf{x}) \frac{\partial \mathcal{L}(\hat{u}(t^{\text{ei},\alpha}))}{\partial \boldsymbol{\boldsymbol{\theta}}}  d\mathbf{x} - \sum_\alpha M^{\text{epi},\alpha} \frac{\partial \mathcal{L}(\hat{u}(\mathbf{x}^{\text{epi},\alpha},t^{\text{epi},\alpha}))}{\partial \boldsymbol{\boldsymbol{\theta}}}\right] 
    \\\nonumber
    & 
     \frac{2\lambda^{\text{b}}}{||\partial \Omega || T } \left[ \int_{\partial \Omega} \int_0^T  [\mathcal{B}(\hat{u})-\hat{g}(\mathbf{x},t) ]\frac{\partial \mathcal{B}(\hat{u})}{\partial \boldsymbol{\boldsymbol{\theta}}} dt d\mathbf{x} - \sum_\alpha \int_0^T  M^{\text{bi},\alpha}(t)  \frac{\partial \mathcal{B}(\hat{u}(\mathbf{x}^{\text{bi},\alpha}))}{\partial \boldsymbol{\boldsymbol{\theta}}} dt \right.
    \\\nonumber
    &\left.\qquad\qquad- \sum_\alpha \int_{\partial \Omega} M^{\text{bi},\alpha}(\mathbf{x}) \frac{\partial \mathcal{B}(\hat{u}(t^{\text{bi},\alpha}))}{\partial \boldsymbol{\boldsymbol{\theta}}}  d\mathbf{x} - \sum_\alpha M^{\text{bpi},\alpha} \frac{\partial \mathcal{B}(\hat{u}(\mathbf{x}^{\text{bpi},\alpha},t^{\text{bpi},\alpha}))}{\partial \boldsymbol{\boldsymbol{\theta}}}\right]
    \\ \nonumber
    &+ \frac{2\lambda^{\text{i}}}{||\Omega|| }\left[\int_\Omega  [\hat{u}(t=t_0) - \hat{h}(\mathbf{x})] \frac{\partial \hat{u}(t=t_0)}{\partial \boldsymbol{\boldsymbol{\theta}}}  d\mathbf{x} -  \sum_\alpha M^{\text{ip},\alpha} \frac{\partial \hat{u}(\mathbf{x}^{\text{ip},\alpha},t_0)}{\partial \boldsymbol{\boldsymbol{\theta}}}\right] .
\end{align}

\section{Matrix $A$ and vector $\mathbf{b}$ and their derivatives for linear PDE problems}\label{sec:coefficients}

We consider the case when the operators $\mathcal{L}$ and $\mathcal{B}$  and the functions $f$, $g$, and $h$ are linear in $\mathbf{v}$. We define the auxiliary vector $\mathbf{v}' = [1,\mathbf{v}^\mathsf{T}]^\mathsf{T}$ , such that $\mathcal{L} = \sum_{\alpha'} v'_{\alpha'}\mathcal{L}^{\alpha'}$, $\mathcal{B} = \sum_{\alpha'} v'_{\alpha'}\mathcal{B}^{\alpha'}$, $f = \sum_{\alpha'} v'_{\alpha'}\hat{f}^{\alpha'}$, $M^{\text{ep}} = \sum_{\alpha'} v'_{\alpha'}M^{\text{ep},\alpha'}$,  $g= \sum_{\alpha'} v'_{\alpha'}g^{\alpha'}$, and $U = \sum_{\alpha'} v'_{\alpha'}U^{\alpha'}$.  The matrix and vector of the linear system (Eq. \eqref{eq:linear_weak_form}) can be written as
\begin{align}
    A_{kl} = \sum_{\alpha'}  v'_{\alpha'}\sum_{\beta'} v'_{\beta'}&\left(\frac{\lambda^{\text{e}} \int_\Omega \int_0^T   \mathcal{L}^{\alpha'}(\phi^k) \mathcal{L}^{\beta'}(\phi^l) dt d\mathbf{x}}{||\Omega|| T } \right.
    \\\nonumber
    &\quad\left.+\frac{\lambda^{\text{b}}\int_{\partial \Omega} \int_0^T  \mathcal{B}^{\alpha'}(\phi^k)  \mathcal{B}^{\beta'}(\phi^l) dt d\mathbf{x}}{||\partial \Omega || T } \right)+ \frac{\lambda^{\text{i}} \int_\Omega  \phi_{0}^k \phi_{0}^l  d\mathbf{x}}{||\Omega|| },
    \\
    b_{k} =\sum_{\alpha'}  v'_{\alpha'}& \left(\sum_{\beta'} v'_{\beta'} \left(\frac{\lambda^{\text{e}} \left[ \int_{\Omega} \int_0^T   \hat{f}^{\alpha'} \mathcal{L}^{\beta'}(\phi^k) dt d\mathbf{x}+\int_0^T   M^{\text{ep},\alpha'} \mathcal{L}^{\beta'}(\phi^k_{\text{ep}}) dt\right] }{||\Omega || T } \right.\right.
    \\\nonumber
    &\left.\left.+\frac{\lambda^{\text{b}} \int_{\partial \Omega} \int_0^T   g^{\alpha'} \mathcal{B}^{\beta'}(\phi^k) dt d\mathbf{x}}{||\partial \Omega || T }\right) + \frac{\lambda^{\text{i}}  \int_\Omega   U^{\alpha'} \phi_{0}^k d\mathbf{x} }{||\Omega|| }\right).
\end{align}

Their gradients are
\begin{align}
   \frac{\partial A_{kl}}{\partial v_{\alpha}}  = \sum_{\alpha}\sum_{\beta'} v'_{\beta'}&\left(\frac{\lambda^{\text{e}} \left[\int_\Omega \int_0^T \left(  \mathcal{L}^{\alpha}(\phi^k) \mathcal{L}^{\beta'}(\phi^l)+\mathcal{L}^{\beta'}(\phi^k) \mathcal{L}^{\alpha}(\phi^l) \right) dt d\mathbf{x}\right]}{||\Omega|| T }  \right.
\\\nonumber
&\quad\left.+ \frac{\lambda^{\text{b}} \left[\int_{\partial \Omega} \int_0^T  \left(\mathcal{B}^{\alpha}(\phi^k)  \mathcal{B}^{\beta'}(\phi^l) +  \mathcal{B}^{\beta'}(\phi^k)  \mathcal{B}^{\alpha}(\phi^l)\right) dt d\mathbf{x}\right]}{||\partial \Omega || T } \right),
    \\
     \frac{\partial b_{k}}{\partial v_{\alpha}} =\sum_{\alpha}&\left(\sum_{\beta'} v'_{\beta'} \left(\frac{\lambda^{\text{e}} \left[\int_{\Omega} \int_0^T   \left(\hat{f}^{\alpha} \mathcal{L}^{\beta'}(\phi^k)+\hat{f}^{\beta'} \mathcal{L}^{\alpha}(\phi^k)\right) dt d\mathbf{x}\right]}{||\Omega || T } \right.\right.
     \\\nonumber
     &+\frac{\lambda^{\text{e}} \left[ \int_0^T   \left(M^{\text{ep},\alpha} \mathcal{L}^{\beta'}(\phi^k)+M^{\text{ep},\beta'} \mathcal{L}^{\alpha}(\phi^k)\right) dt \right]}{||\Omega || T } 
    \\\nonumber
    &\left.\left.+\frac{\lambda^{\text{b}}  \left[\int_{\partial \Omega} \int_0^T   \left(g^{\alpha} \mathcal{B}^{\beta'}(\phi^k) + g^{\beta'} \mathcal{B}^{\alpha}(\phi^k) \right) dt d\mathbf{x}\right]}{||\partial \Omega || T }\right) + \frac{\lambda^{\text{i}}  \int_\Omega   U^{\alpha} \phi_{0}^k d\mathbf{x} }{||\Omega|| }\right),
\end{align}
where $\sum_{\alpha}$ denotes the summation over the subindices of $\mathbf{v}$ instead of $\mathbf{v}'$. The Hessians are
\begin{align}
   \frac{\partial^2 A_{kl}}{\partial v_{\alpha} \partial v_{\beta}}  = \sum_{\alpha}\sum_{\beta}& \left(\frac{\lambda^{\text{e}} \left[\int_\Omega \int_0^T \left(  \mathcal{L}^{\alpha}(\phi^k) \mathcal{L}^{\beta}(\phi^l)+\mathcal{L}^{\beta}(\phi^k) \mathcal{L}^{\alpha}(\phi^l) \right) dt d\mathbf{x}\right]}{||\Omega|| T }  \right.
\\\nonumber
&\left.\quad+ \frac{\lambda^{\text{b}} \left[\int_{\partial \Omega} \int_0^T  \left(\mathcal{B}^{\alpha}(\phi^k)  \mathcal{B}^{\beta}(\phi^l) +  \mathcal{B}^{\beta}(\phi^k)  \mathcal{B}^{\alpha}(\phi^l)\right) dt d\mathbf{x}\right]}{||\partial \Omega || T } \right),
    \\
     \frac{\partial^2 b_{k}}{\partial v_{\alpha} \partial v_{\beta}} =\sum_{\alpha}\sum_{\beta}&\left(\frac{\lambda^{\text{e}}  \left[\int_{\Omega} \int_0^T   \left(\hat{f}^{\alpha} \mathcal{L}^{\beta}(\phi^k)+\hat{f}^{\beta} \mathcal{L}^{\alpha}(\phi^k)\right) dt d\mathbf{x}\right]}{||\Omega || T } \right.
\\\nonumber
&\quad+\frac{\lambda^{\text{e}}  \left[\int_0^T   \left(M^{\text{ep},\alpha} \mathcal{L}^{\beta}(\phi^k)+M^{\text{ep},\beta} \mathcal{L}^{\alpha}(\phi^k)\right) dt \right]}{||\Omega || T }
\\\nonumber
&\left.\quad+\frac{\lambda^{\text{b}} \sum_{\alpha}\sum_{\beta}  \left[\int_{\partial \Omega} \int_0^T   \left(g^{\alpha} \mathcal{B}^{\beta}(\phi^k) + g^{\beta} \mathcal{B}^{\alpha}(\phi^k) \right) dt d\mathbf{x}\right]}{||\partial \Omega || T } \right).
\end{align}

\section{Analytical solution to the diffusion equation with local forcing} \label{app:analytical_PS}

\sloppy To solve the problem described by Section \ref{sec:prob_gwflow}, we use the Green's functions solution $u(x,t)=\int_0^{\infty} \int_0^1 G(x,s,t,\xi) f(s,\xi) ds d\xi$.

Since the operator $G$ is translation invariant (because the equation has constant coefficients), we can write
\begin{equation}
    u(x,t)= Q \int_0^{\infty} \int_0^1 G(x-s,t-\xi) \delta(s-x^{\text{ep}}) ds \;d\xi =  Q \int_0^{\infty} G(x-x^{\text{ep}},t-\xi)d\xi,
\end{equation}
which holds as long as $0<x^{\text{ep}}<1$. The Green's function for the operator $\mathcal{L}=\frac{\partial }{\partial t} - k \frac{\partial^2}{\partial x^2}$ is 
\begin{align}
    G=\Theta(t)\sqrt{\frac{1}{4\pi k t}} \exp{\left( -\frac{x^2}{4kt} \right)} + C,
\end{align}
where $\Theta(t)$ is the Heaviside step function. Then we can rewrite
\begin{equation}\label{eq:Greens-integral}
    h(x,t)= Q \int_0^t \sqrt{\frac{1}{4\pi k (t-\xi)}} \exp{\left( -\frac{(x-x^{\text{ep}})^2}{4k(t-\xi)} \right)}  \;d\xi.
\end{equation}
We know the solution to the following integral
\begin{align}
    \int_0^u \sqrt{\frac{1}{u'}} \exp{\left( -\frac{1}{u'} \right)}  du' = 2  \exp{\left( -\frac{1}{u} \right)} \sqrt{u} + 2 \sqrt{\pi} \left(\mathrm{erf}\left( \frac{1}{\sqrt{u}} \right) - 1 \right)
\end{align}
which we use to perform the substitution $u'(\xi)=\frac{4k(t-\xi)}{(x-x^{\text{ep}})^2}$, $du'=-\frac{4k}{(x-x^{\text{ep}})^2} d\xi$, $u'(0)=\frac{4kt}{(x-x^{\text{ep}})^2}$, $u'(t)=0$, to obtain
\begin{align}
    h(x,t)= \frac{Q}{2k} |x-x^{\text{ep}}| \mathrm{erfc}\left( \frac{|x-x^{\text{ep}}|}{\sqrt{4kt}} \right) -Q \sqrt{\frac{t}{k\pi}} \exp{\left( -\frac{(x-x^{\text{ep}})^2}{4kt} \right)}
\end{align}

This doesn't satisfy the boundary conditions. Instead, it satisfies $h(x=-\infty,t)=h(x=\infty,t)=h(x,t=0)=0$.

\subsection{Method of images}

To satisfy the boundary condition at points different from infinity, we can use the method of images, given that the equation is linear. We write the solution as the sum of different individual solutions that don't satisfy the boundary conditions
\begin{equation}
    u(x,t) = \sum_i a_i \left(\frac{Q}{2k} |x-{x_i^{\text{ep}}}'| \mathrm{erfc}\left(\frac{|x-{x_i^{\text{ep}}}'|}{\sqrt{4kt}} \right) -Q \sqrt{\frac{t}{k\pi}} \exp{\left( -\frac{(x-{x_i^{\text{ep}}}')^2}{4kt} \right)}\right),
\end{equation}
with the condition that the only ${x_i^{\text{ep}}}'$ in the interval $x \in [0,1]$ is $x^{\text{ep}}$. We need to satisfy $h(0,t) =  h(1,t) =0$. Adding a constant solution to satisfy the boundary conditions we obtain
\begin{align}
    u(x,t) =& \sum_{n=-\infty}^{\infty} \left(\frac{Q}{2k} |\alpha_+| \mathrm{erfc}\left(\frac{|\alpha_+|}{\sqrt{4kt}} \right) -Q \sqrt{\frac{t}{k\pi}} \exp{\left( -\frac{(\alpha_+)^2}{4kt} \right)} \right.
    \\\nonumber
    & \left. - \frac{Q}{2k} |\alpha_-| \mathrm{erfc}\left(\frac{|\alpha_-|}{\sqrt{4kt}} \right) +Q \sqrt{\frac{t}{k\pi}} \exp{\left( -\frac{(\alpha_-)^2}{4kt} \right)}\right) + 1,
\end{align}
with $\alpha_{\pm} = x-(2n\pm x^{\text{ep}})$.

\section{Neural Tangent Kernel analysis of PINN solutions to PDEs with local forcing} \label{app:NTK}

We analyze the behavior of the losses of a PDE system with point source forcing using Neural Tangent Kernel (NTK) theory \cite{Wang2022NTK}, with $\hat{f}=0$ and $M^{\text{ep}}=1$. We define $\mathbf{r}^{\text{e}}$, $\mathbf{r}^{\text{ep}}$,$\mathbf{r}^{\text{b}}$, and $\mathbf{r}^{\text{i}}$, the vectors of the PDE, Dirac delta source, boundary, and initial residual values at the quadrature points with components 
\begin{equation}
    r^{\text{e}}_i =  \sqrt{\frac{\lambda^{\text{e}}}{N^{\text{e}}}}  \mathcal{L}(\hat{u}^{\text{e}}_i),\quad r^{\text{ep}}_i= \frac{\lambda^{\text{e}}}{N^{\text{ep}}}  \mathcal{L}(\hat{u}^{\text{ep}}_i),\quad r_i^{\text{b}} = \sqrt{\frac{\lambda^{\text{b}}}{N^{\text{b}}}} [\mathcal{B}(\hat{u}^{\text{b}}_i) -g_{i}],\quad r_i^{\text{i}} = \sqrt{\frac{\lambda^{\text{i}}}{N^{\text{i}}}} [\hat{u}^{\text{i}}_i -h_{i}],
\end{equation}
of sizes $N_{\Omega\times T}^\text{e}$, $N_{T}^\text{ep}$, $N_{\Omega\times T}^\text{b}$, and $N_{\Omega}^\text{i}$, respectively. The total loss function as defined in \eqref{eq:DLLR-derivatives} can be written as 
\begin{equation}
    \mathcal{R}(\boldsymbol{\theta})=\mathbf{r}^{\text{e}}\cdot \mathbf{r}^{\text{e}}- 2 \;\mathbf{1}_{N^{\text{ep}}}\cdot\mathbf{r}^{\text{ep}} +  \mathbf{r}^{\text{b}}\cdot \mathbf{r}^{\text{b}} + \mathbf{r}^{\text{i}} \cdot \mathbf{r}^{\text{i}},
\end{equation}
where $\mathbf{1}_{N^{\text{ep}}}$ is a vector of ones of size $N_{T}^\text{ep}$. The minimization by gradient descent of the loss function leads to the continuous-time gradient flow system (with the training time variable defined as $\tau$)
\begin{equation}
    \frac{1}{2}\frac{d\boldsymbol{\theta}}{d\tau}=-\frac{1}{2}\nabla_{\boldsymbol{\theta}}\mathcal{R} (\boldsymbol{\theta}) = \mathbf{r}^{\text{e}}\cdot \frac{\partial \mathbf{r}^{\text{e}}}{\partial \boldsymbol{\theta}} -  \mathbf{1}_{N^{\text{ep}}}\cdot  \frac{\partial \mathbf{r}^{\text{ep}}}{\partial \boldsymbol{\theta}}+  \mathbf{r}^{\text{b}}\cdot \frac{\partial \mathbf{r}^{\text{b}}}{\partial \boldsymbol{\theta}}  + \mathbf{r}^{\text{i}}\cdot \frac{\partial \mathbf{r}^{\text{i}}}{\partial \boldsymbol{\theta}} .
\end{equation}
For any residual $\mathbf{r}_{\alpha}$ with $\alpha \in [\text{e},\text{ep},\text{b},\text{i}]$, the system that defines its evolution is given by the chain rule as $\frac{d \mathbf{r}_{\alpha}}{d\tau}=\frac{\partial}{\partial\theta} \frac{d\boldsymbol{\theta}}{d\tau}$, leading to a dynamical system for the residuals
\begin{equation}
   \frac{d}{d\tau} \begin{bmatrix}
    \mathbf{r}^{\text{e}}\\
    \mathbf{r}^{\text{ep}}\\
    \mathbf{r}^{\text{b}}\\
    \mathbf{r}^{\text{i}}
    \end{bmatrix} 
     = -2
    \begin{bmatrix}
        K^{\text{e},\text{e}}(\tau) & K^{\text{e},\text{ep}}(\tau) & K^{\text{e},\text{b}}(\tau) & K^{\text{e},\text{i}}(\tau)\\
        K^{\text{ep},\text{e}}(\tau) & K^{\text{ep},\text{ep}}(\tau) & K^{\text{ep},\text{b}}(\tau) & K^{\text{ep},\text{i}}(\tau)\\
        K^{\text{b},\text{e}}(\tau) & K^{\text{b},\text{ep}}(\tau) & K^{\text{b},\text{b}}(\tau) & K^{\text{b},\text{i}}(\tau)\\
        K^{\text{i},\text{e}}(\tau) & K^{\text{i},\text{ep}}(\tau) & K^{\text{i},\text{b}}(\tau) & K^{\text{i},\text{i}}(\tau)
    \end{bmatrix}
    \begin{bmatrix}
        \mathbf{r}^{\text{e}}\\
        -\mathbf{1}_{N^{\text{ep}}}\\
        \mathbf{r}^{\text{b}}\\
        \mathbf{r}^{\text{i}}
    \end{bmatrix},
\end{equation}
where $(K^{\alpha,\beta})_{i,j} (\tau)= \mathrm{Ker}^{\alpha,\beta}(\mathbf{x}_i,t_i,\mathbf{x}_j,t_j,\tau)= \sum_k^{N_{\theta}} \frac{\partial r^{i}_\alpha(\tau) }{\partial \theta_k} \frac{\partial r^{j}_\beta(\tau)}{\partial \theta_k}$ with $\alpha,\beta \in [\text{e},\text{ep},\text{b},\text{i}]$, and $\mathrm{Ker}$ is the Neural Tangent Kernel. 

As the width of the Neural Network increases, the Neural Tangent Kernel tends to a limit function that doesn't evolve in time \cite{jacot2018neural,Wang2022NTK}. An additional condition for this is that any integral over training time of the magnitude of the derivative of the loss function with respect to the outputs of the neural network must remain bounded \cite{jacot2018neural}. Under this condition, the dynamical system becomes linear, and it has an analytical solution. The solution for $\mathbf{r}^{\text{e}}$, $\mathbf{r}^{\text{b}}$, and $\mathbf{r}^{\text{i}}$ is independent of $\mathbf{r}^{\text{ep}}$
\begin{equation}
    \begin{bmatrix}
    \mathbf{r}^{\text{e}} (\tau)\\
    \mathbf{r}^{\text{b}} (\tau)\\
    \mathbf{r}^{\text{i}} (\tau)
    \end{bmatrix} 
    =    \begin{bmatrix}
    \mathbf{r}^{\text{e}}_*\\
    \mathbf{r}^{\text{b}}_*\\
    \mathbf{r}^{\text{i}}_*
    \end{bmatrix}  
    + \exp\left(-2\begin{bmatrix}
        K^{\text{e},\text{e}} & K^{\text{e},\text{b}} & K^{\text{e},\text{i}}\\
        K^{\text{b},\text{e}} & K^{\text{b},\text{b}} & K^{\text{b},\text{i}}\\
        K^{\text{i},\text{e}}  & K^{\text{i},\text{b}} & K^{\text{i},\text{i}}
    \end{bmatrix}  \tau\right) 
    \left(
    \begin{bmatrix}
    \mathbf{r}^{\text{e}} (0)\\
    \mathbf{r}^{\text{b}} (0)\\
    \mathbf{r}^{\text{i}} (0)
    \end{bmatrix} 
    -\begin{bmatrix}
    \mathbf{r}^{\text{e}}_*\\
    \mathbf{r}^{\text{b}}_*\\
    \mathbf{r}^{\text{i}}_*
    \end{bmatrix} 
    \right),
\end{equation}
where
\begin{equation}\label{eq:limit_residual}
    \begin{bmatrix}
        K^{\text{e},\text{e}} & K^{\text{e},\text{b}} & K^{\text{e},\text{i}}\\
        K^{\text{b},\text{e}} & K^{\text{b},\text{b}} & K^{\text{b},\text{i}}\\
        K^{\text{i},\text{e}}  & K^{\text{i},\text{b}} & K^{\text{i},\text{i}}
    \end{bmatrix} \begin{bmatrix}
    \mathbf{r}^{\text{e}}_*\\
    \mathbf{r}^{\text{b}}_*\\
    \mathbf{r}^{\text{i}}_*
    \end{bmatrix}  = 
    \begin{bmatrix}
        K^{\text{e},\text{ep}} \\
        K^{\text{b},\text{ep}}\\
        K^{\text{i},\text{ep}}
    \end{bmatrix} \mathbf{1}_{N^{\text{ep}}}.
\end{equation}
Then, we can obtain the solution for $\mathbf{r}^{\text{ep}}$
\begin{align}
    \mathbf{r}^{\text{ep}} (\tau) =& \mathbf{r}^{\text{ep}} (0) -2\tau \left(
    [K^{\text{ep},\text{e}} , K^{\text{ep},\text{b}} , K^{\text{ep},\text{i}}] 
    \begin{bmatrix}
    \mathbf{r}^{\text{e}}_*\\
    \mathbf{r}^{\text{b}}_*\\
    \mathbf{r}^{\text{i}}_*
    \end{bmatrix} + K^{\text{ep},\text{ep}} \mathbf{1}_{N^{\text{ep}}} \right) 
    \\\nonumber
    &-  \begin{bmatrix}
    K^{\text{ep},\text{e}} \\ K^{\text{ep},\text{b}} \\ K^{\text{ep},\text{i}}
    \end{bmatrix} ^\mathsf{T} 
    \begin{bmatrix}
        K^{\text{e},\text{e}} & K^{\text{e},\text{b}} & K^{\text{e},\text{i}}\\
        K^{\text{b},\text{e}} & K^{\text{b},\text{b}} & K^{\text{b},\text{i}}\\
        K^{\text{i},\text{e}}  & K^{\text{i},\text{b}} & K^{\text{i},\text{i}}
    \end{bmatrix} ^{-1}
    \\\nonumber
    &\quad \left(I-\exp\left(-2\begin{bmatrix}
        K^{\text{e},\text{e}} & K^{\text{e},\text{b}} & K^{\text{e},\text{i}}\\
        K^{\text{b},\text{e}} & K^{\text{b},\text{b}} & K^{\text{b},\text{i}}\\
        K^{\text{i},\text{e}}  & K^{\text{i},\text{b}} & K^{\text{i},\text{i}}
    \end{bmatrix} \tau\right) \right) 
    \left(
    \begin{bmatrix}
    \mathbf{r}^{\text{e}} (0)\\
    \mathbf{r}^{\text{b}} (0)\\
    \mathbf{r}^{\text{i}} (0)
    \end{bmatrix} 
    -\begin{bmatrix}
    \mathbf{r}^{\text{e}}_*\\
    \mathbf{r}^{\text{b}}_*\\
    \mathbf{r}^{\text{i}}_*
    \end{bmatrix} 
    \right)
\end{align}

Taking into account that all the matrices are positive semi definite, the linear solutions show exponential decay, and we have $\lim_{\tau\to\infty} \mathbf{r} (\tau) = \mathbf{r}_*$, with $\mathbf{r} = [{\mathbf{r}^{\text{e}}}^{\mathsf{T}},
    {\mathbf{r}^{\text{b}}}^{\mathsf{T}},
    {\mathbf{r}^{\text{i}}}^{\mathsf{T}}]^{\mathsf{T}}$ and $\mathbf{r}_* = [{\mathbf{r}^{\text{e}}_*}^{\mathsf{T}},
    {\mathbf{r}^{\text{b}}_*}^{\mathsf{T}},
    {\mathbf{r}^{\text{i}}_*}^{\mathsf{T}}]^{\mathsf{T}}$, and $\lim_{\tau\to\infty} \frac{\mathbf{r}^{\text{ep}}(\tau)}{\tau} = -2\mathbf{r}^{\text{ep}}_*$, with $\mathbf{r}^{\text{ep}}_*=
   [K^{\text{ep},\text{e}} , K^{\text{ep},\text{b}} , K^{\text{ep},\text{i}}] {\mathbf{r}_*}^{\mathsf{T}}+ K^{\text{ep},\text{ep}} \mathbf{1}_{N^{\text{ep}}}$. This is in contrast to residuals converging to zero in typical PINN problems \cite{Wang2022NTK}. Since the Neural Tangent Kernel tends to a limit kernel (in general with nonzero practical radius of influence) as the width of the Neural Network becomes infinite \cite{jacot2018neural}, the cross terms $K^{\text{e},\text{ep}}$, $K^{\text{ep},\text{b}}$, and $K^{\text{ep},\text{i}}$ will be non-zero in general, and so will be $\mathbf{r}_*$ (from Eq. \eqref{eq:limit_residual}).

On the other hand, the Neural Tangent Kernel for an RBF network has a practical radius of influence proportional to $h$. For the particular case of a time independent problem, with Gaussian RBFs, applying an identity operator, and assuming boundary effects are negligible, the limit kernel for infinite terms in the RBF expansion is 
\begin{align}
    \lim_{\Delta\to0}\mathrm{Ker} (\mathbf{x}_i,t_i,\mathbf{x}_j,t_j)&=\lim_{\Delta\to0}\sum_{k=1}^N W(||\mathbf{x}_i-\mathbf{x}^c_k||,h)W(||\mathbf{x}_j-\mathbf{x}^c_k||,h)
    \\\nonumber
    &\propto\int_{\Omega} W(||\mathbf{x}_i-\mathbf{x}^c_k||,h)W(||\mathbf{x}_j-\mathbf{x}^c_k||,h) d\mathbf{x}^c_k
    \propto W(||\mathbf{x}_i-\mathbf{x}_j||,\sqrt{2}h).
\end{align}
The practical radius of influence of the kernel can be made tend to zero by making $h\to0$ as $\Delta\to 0$ (though $h$ must converge slower than $\Delta$ due to terms decreasing with $\frac{\Delta}{h}$, Section \ref{sec:res_ADE-inst}), making the cross terms $K^{\text{e},\text{ep}}$, $K^{\text{ep},\text{b}}$, and $K^{\text{ep},\text{i}}$ also zero. The absence of forcing terms in the infinite size limit means the residuals can be reduced to arbitrarily small values, $\lim_{\tau\to\infty} \mathbf{r} (\tau) =0$.